\documentclass[acmsmall]{acmart}
\AtBeginDocument{%
  }

\usepackage{pifont}
\usepackage{colortbl}
\usepackage{multirow}
\usepackage{url}

\usepackage{amsmath,amssymb,amsfonts, bm}
\usepackage{colortbl} 
\usepackage{xcolor}
\usepackage{array}  
\usepackage{subfigure}
\setcopyright{acmcopyright}
\usepackage{twemojis}

\usepackage[ruled,linesnumbered]{algorithm2e}
\usepackage[export]{adjustbox}
\usepackage{multicol}
\usepackage{enumitem}
\usepackage{titletoc}
\usepackage{tcolorbox}
\usepackage{hyperref}

\setcopyright{acmlicensed}
\acmJournal{POMACS}
\acmYear{2025} \acmVolume{9} \acmNumber{3} \acmArticle{60} \acmMonth{12}\acmDOI{10.1145/3771575}

\begin{document}

\title{Optimizing Storage Overhead of User Behavior Log for ML-embedded Mobile Apps}

\author{Chen Gong}
\email{gongchen@sjtu.edu.cn}
\orcid{0000-0003-0333-6418}
\affiliation{
    \institution{Shanghai Jiao Tong University}
    \city{Shanghai}
    \country{China}
}

\author{Yan Zhuang}
\email{zhuang00@sjtu.edu.cn}
\orcid{0009-0006-0702-5093}
\affiliation{
    \institution{Shanghai Jiao Tong University}
    \city{Shanghai}
    \country{China}
}

\author{Zhenzhe Zheng}
\authornote{Zhenzhe Zheng is the corresponding author.}
\email{zhengzhenzhe@sjtu.edu.cn}
\orcid{0000-0002-5094-5331}
\affiliation{
    \institution{Shanghai Jiao Tong University}
    \city{Shanghai}
    \country{China}
}

\author{Yiliu Chen}
\email{chenyiliu@bytedance.com}
\orcid{0009-0002-2775-3566}
\affiliation{
    \institution{ByteDance}
    \city{Hangzhou}
    \country{China}
}

\author{Sheng Wang}
\email{wangsheng.john@bytedance.com}
\orcid{0009-0009-5554-6207}
\affiliation{
    \institution{ByteDance}
    \city{Hangzhou}
    \country{China}
}

\author{Fan Wu}
\email{fwu@cs.sjtu.edu.cn}
\orcid{0000-0003-0965-9058}
\affiliation{
    \institution{Shanghai Jiao Tong University}
    \city{Shanghai}
    \country{China}
}

\author{Guihai Chen}
\email{gchen@cs.sjtu.edu.cn}
\orcid{0000-0002-6934-1685}
\affiliation{
    \institution{Shanghai Jiao Tong University}
    \city{Shanghai}
    \country{China}
}


\begin{abstract}
Machine learning (ML) models are increasingly integrated into modern mobile apps to enable personalized and intelligent services. These models typically rely on rich input features derived from historical user behaviors to capture user intents.
However, as ML-driven services become more prevalent, recording necessary user behavior data imposes substantial storage cost on mobile apps, leading to lower system responsiveness and more app uninstalls.
To address this storage bottleneck, we present {\tt AdaLog}, a lightweight and adaptive system designed to improve the storage efficiency of user behavior log in ML-embedded mobile apps, without compromising model inference accuracy or latency.
We identify two key inefficiencies in current industrial practices of user behavior log: 
(i) redundant logging of overlapping behavior data across different features and models, and 
(ii) sparse storage caused by storing behaviors with heterogeneous attribute descriptions in a single log file. 
To solve these issues, {\tt AdaLog} first formulates the elimination of feature-level redundant data as a maximum weighted matching problem in hypergraphs, and proposes a hierarchical algorithm for efficient on-device deployment. 
Then, {\tt AdaLog} employs a virtually hashed attribute design to distribute heterogeneous behaviors into a few log files with physically dense storage. 
Finally, to ensure scalability to dynamic user behavior patterns, {\tt AdaLog} designs an incremental update mechanism to minimize the I/O operations needed for adapting outdated behavior log.
We implement a prototype of {\tt AdaLog} and deploy it into popular mobile apps in collaboration with our industry partner. Evaluations on real-world user data show that {\tt AdaLog} reduces behavior log size by $19\%$ to $44\%$ with minimal system overhead (only 2 seconds latency and 15 MB memory usage), providing a more efficient data foundation for broader adoption of on-device ML.
\end{abstract}



\received{July 2025}
\received[revised]{September 2025}
\received[accepted]{October 2025}

\maketitle

\section{Introduction}
\label{sec: introduction}
The rapid evolution of smartphones has driven the widespread integration of machine learning (ML) models into modern mobile apps~\cite{sarker2021mobile, liu2016lasagna, DBLP:conf/imc/AlmeidaLMDLL21, DBLP:conf/www/XuLLLLL19, gong2024delta, gong2025two}, empowering a new era of context-aware and personalized app services.
Representative examples include 
personalized product recommendations in e-commerce platforms~\cite{han2021deeprec, gharibshah2021user, cheng2016wide, covington2016deep}, search results ranking in search engines~\cite{karmaker2017application, ziakis2019important, zhang2018towards}, as well as video preloading and bandwidth management in multimedia streaming apps~\cite{yeo2018neural, mehrabi2018edge, tran2018adaptive}.
Unlike traditional vision or language models that use \textit{static} input features (\textit{e.g.,} pixel values or token embeddings), ML models deployed in mobile apps rely on \textit{dynamic} input features extracted from evolving user behaviors to capture dynamic contexts and user intents~\cite{lai2023adaembed, lindorfer2015marvin, ouyang2022learning} (\textit{e.g.,} the recent product clicks and video views could partially reflect a user's shifting preference). 

To enable accurate feature computation for on-device model inferences, current mobile apps store all necessary user behavior data in a dedicated file known as the \textit{user behavior log}, which acts as a critical middle layer to connect physical user behaviors and on-device ML models~(\S\ref{sec: role of behavior log}).
Specifically, every user interaction with the app's graphical user interface (GUI) can be captured as a structured \textit{behavior event}, which consists of numerous different \textit{attributes} to provide a rich, detailed description of each user behavior. 
When a behavior event occurs in real time, different model features apply their own \textit{filters} to check if the current event is relevant, pick out necessary attribute subsets and store them as event rows in behavior log, which is typically implemented as SQLite databases for both Android and iOS devices~\cite{gaffney2022sqlite, kreibich2010using, oh2015sqlite}.
To facilitate the fast retrieval of relevant events for feature computation, behavior log also maintains an index structure to map each feature to the physical storage addresses of its relevant event rows within the database.

\textbf{Storage Bottleneck.}
Despite its essential role, behavior log introduces an overlooked storage bottleneck that limits the scalability and broader adoption of ML-powered mobile services (\S\ref{sec: storage bottleneck}). 
Through our empirical study of over 20 ML models deployed in real-world mobile apps, spanning domains of live streaming, e-commerce, searching and advertising, we observe that behavior log size increases dramatically with the growing number and scale of on-device ML models, consuming up to 50\% of the app size. 
Public statistics~\cite{app_uninstall} show that excessive app size is the primary cause of app uninstalls, and our industrial statistics also reveal that each additional 10 MB in app size of TikTok can lead to around 61,000 fewer daily active users and \$7000 financial losses per day. Consequently, the storage overhead of behavior log has emerged as a practical obstacle to the widespread adoption of ML-powered mobile services.

\textbf{Our Motivation.}
In this work, we aim to tackle a crucial but unexplored problem: optimizing storage efficiency of behavior log for ML-embedded mobile apps. The optimization opportunities stem from our two key observations of real-world mobile data~(\S\ref{sec: optimization opportunities}).
First, many features of ML models rely on partially overlapping attributes derived from the same behavior event, but are recorded as separate event rows in behavior log for fast retrieval, resulting in redundant data storage.
Second, different behavior events are described by heterogeneous attributes but are stored in a unified log file, which leads to massive null values for each event's irrelevant attributes and results in sparse storage.
These inefficiencies reveal promising opportunities for storage optimization of behavior logs.

To this end, we propose {\tt AdaLog}, the first system designed to reduce the storage cost of behavior log without sacrificing on-device inference accuracy or latency. {\tt AdaLog} is built on two core insights: 
(i) Reduce data storage redundancy by merging event rows that are logged by different features but come from the same behavior events; 
(ii) Eliminate storage sparsity by distributing behavior events with heterogeneous attribute sets into separate sub-log files for dense storage.
By tacking these storage-level inefficiencies, {\tt AdaLog} provides a more powerful data foundation for ML-embedded mobile apps and offers a new optimization dimension that complements existing works on resource-efficient model inference, as they primarily focused on memory, computation, energy, etc~(\S\ref{sec: Related Work}).

\textbf{Challenges.}
Designing and implementing {\tt AdaLog} system involves three key challenges.

\textit{First, identifying which features to merge their event rows for redundancy elimination is an NP-hard problem.} 
While merging redundant event rows across any features can reduce data storage cost, it introduces additional overhead for the index structure, as more physical addresses have to be tracked to distinguish each feature's relevant event rows from the merged ones~(\S\ref{sec: feature-level data merging}). 
We show that deriving the optimal feature-level data merging strategy is equivalent to solving the maximum weighted matching problem in a hypergraph: each node denotes a feature and each hyperedge connecting multiple nodes denotes a candidate feature group, weighted by the potential storage savings after data merging. This optimization problem is NP-hard and computationally expensive.

\textit{Second, deciding where to store each behavior's event rows to eliminate storage sparsity is not straightforward either. } 
Ideally, different user behaviors described by distinct attribute sets should be stored in separate log files, ensuring that event rows in one file share the same attribute columns.
However, user behaviors captured by mobile apps are massive and highly diverse, leading to hundreds of small, fragmented log files~(\S\ref{sec: behavior-level log splitting}). 
Managing such a fragmented storage system incurs substantial metadata overhead (e.g., tracking table names, attribute column formats, file sizes, etc), making it difficult to simultaneously achieve low storage sparsity and high storage efficiency.

\textit{Third, maintaining an up-to-date behavior log for dynamic user behavior patterns is essential but costly.}
As behaviors of mobile users are unpredictable and evolving over time, the optimal strategies of feature-level data merging and behavior-level log splitting are also dynamic, requiring frequent re-optimization to preserve storage efficiency. Unfortunately, applying the strategy changes typically necessitates the reconstruction of behavior logs, involving large-scale I/O operations such as reading, writing and indexing. This imposes serious scalability challenges for deploying the log optimization system in industrial-scale mobile apps. 

\textbf{Our Solutions.}
To address these challenges, {\tt AdaLog} introduces three core techniques to unlock the full optimization potential of behavior log.
First, to identify an effective feature-level data merging strategy without excessive computation, we develop a hierarchical merging algorithm with polynomial time complexity.
Instead of directly solving the hypergraph-based NP-hard problem, we decompose the solving process into multiple iterations. In each iteration, we reduce the hypergraph to a weighted 2D graph and merge only pairs of feature groups, iteratively refining the merging plan.
Second, to achieve both low storage sparsity and metadata overhead, we introduce a virtually hashed attribute naming scheme, which allows heterogeneous attributes of different user behaviors to share the same virtual name and to be stored in one physical column of  the log file. This reduces the required number of log files from the number of behavior types ($\approx\!250$) to the number of possible attribute counts ($\approx\!20$).
Finally, to support efficient adaptation to dynamic user behavior patterns, we develop an incremental update mechanism that avoids full log reconstruction. Observing that changes in behavior patterns typically impact only partial optimization strategies and logged data, we propose a shrink-and-expand method that aligns past and new strategies to reuse as much existing data as possible and incrementally update the affected portions. This design minimizes I/O overhead and enables fast and scalable adaptation for resource-constrained devices.

\textbf{Implementation and Evaluation.}
We have implemented {\tt AdaLog} as a lightweight Python package and evaluate it in industrial mobile apps with the help of our industrial partner~(\S\ref{sec: Evaluation}), involving tens of practical on-device models, hundreds of real-world testing users and four service domains (live streaming, e-commerce, search and advertising). 
Compared to industry-standard behavior log designs, {\tt AdaLog} achieves up to $44\%$ average reduction for behavior log size without compromising inference latency or accuracy, and maintaining minimal system costs of $2$ seconds latency and $15$MB memory footprint.

\textbf{Contributions} of this work are summarized as follows:\\
$\bullet$ We identify a critical but overlooked storage bottleneck in deploying ML models within industrial mobile apps, analyzing its root causes and optimization opportunities.\\
$\bullet$ We design and implement the first storage optimization system for behavior logs in ML-embedded mobile apps, which reduces the storage redundancy and sparsity without sacrificing on-device model inference accuracy or latency, providing a more powerful data foundation for on-device ML.\\
$\bullet$ We evaluate {\tt AdaLog} on industrial mobile apps and real-world users, demonstrating significant storage savings and superior system efficiency compared to industry baselines.

\section{Background and Motivation}
\label{sec: background}
In this section, we first elaborate on the critical role and workflow of behavior log in supporting ML-embedded mobile apps~(\S\ref{sec: role of behavior log}). 
Next, we analyze the storage bottleneck of behavior log based on statistics from industrial mobile apps~(\S\ref{sec: storage bottleneck}). 
Finally, we identify key optimization opportunities 
	and corresponding design choices that motivate our system
	~(\S\ref{sec: optimization opportunities}).

\subsection{Role and Workflow of Behavior Log in Mobile Apps}
\label{sec: role of behavior log}
ML models have become a core component of modern mobile apps~\cite{sarker2021mobile, liu2016lasagna, DBLP:conf/imc/AlmeidaLMDLL21, DBLP:conf/www/XuLLLLL19, gong2024delta, gong2024ode}, powering various intelligent and personalized services by consuming private user data on mobile devices~\cite{covington2016deep, han2021deeprec, gharibshah2021user, karmaker2017application, ziakis2019important, liu2025non, zhang2018towards, liu2025enabling, gomez2015netflix, gong2023store, kamaraju2013novel, yeo2018neural, mehrabi2018edge, tran2018adaptive, liu2025latency}.
For modern mobile apps, ML models practically deployed on mobile devices rely on massive input features derived from a user's various historical behaviors to capture evolving contextual information and user intents~\cite{lai2023adaembed, lindorfer2015marvin, ouyang2022learning}. This necessitates recording relevant behavior data in a specified file called \textit{behavior log}, which is typically implemented as lightweight SQLite databases on both Android and iOS platforms~\cite{obradovic2019performance, feiler2015using, oh2015sqlite}. 

\begin{figure}
    \centering
    \includegraphics[width=\linewidth]{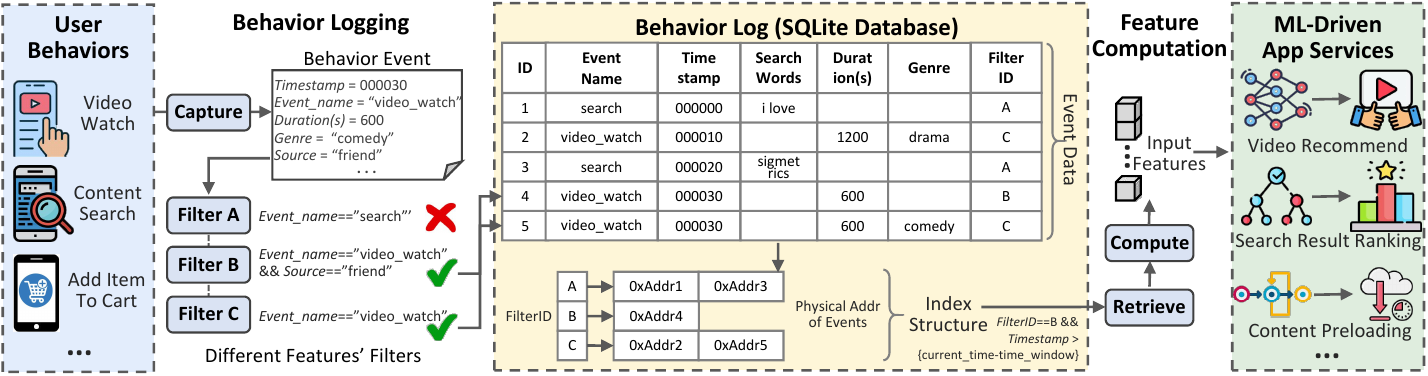}
    \caption{In mobile apps like TikTok, behavior log acts as a middle layer between user behaviors and ML-driven services:   
    	(i) Each user behavior is \textit{captured} as an event containing multiple attributes for description; (ii) Different features apply distinct \textit{filters} to judge event relevance and select necessary attributes to store in behavior log, consisting of event data and index structure; (iii) During real-time feature computation, each feature uses index structure and time window constraints to \textit{retrieve} relevant rows for accurate computation.}
    \Description{In mobile apps like TikTok, behavior log acts as a middle layer between user behaviors and ML-driven services:   
    (i) Each user behavior is \textit{captured} as an event involving multiple attributes for description; (ii) Different features apply distinct \textit{filters} to judge event relevance and select necessary attributes to store in behavior log, comprising event data and index structure; (iii) During real-time feature computation, each feature uses index structure and time window constraints to \textit{retrieve} relevant rows for accurate computation.}
    \label{fig: data processing pipeline}
\end{figure}

The work flow of behavior log is shown in Figure \ref{fig: data processing pipeline}, which serves as a critical middle layer to connect physical user behaviors and on-device ML models through two stages: behavior logging and feature computation. 

\textbf{Behavior Logging.} 
During app usage, each user interaction with the smartphone can be captured as a behavior \textit{event} (\textit{e.g.}, ``video watch''), which is represented as a structured format of numerous \textit{attributes}, including behavior-independent attributes (\textit{e.g.}, event name, timestamp) for identification and behavior-specific attributes to describe the behavior in multiple orthogonal dimensions (\textit{e.g.}, duration, genre, volume for a ``video watch'' event).
However, each input feature is typically used to reflect a certain dimension over a specified context (\textit{e.g.}, the average watch time of videos shared by friends). Thus, in industrial practices, each feature applies a \textit{filter} to 
(i) check whether current behavior is relevant by examining specific attribute values (\textit{e.g.}, ``event name''=``video watch'' and ``source''=``friend''), 
(ii) select necessary attributes from the matching event and append FilterID attribute, (iii) record them as an event row in behavior log. 
To enable low-latency data retrieval for real-time feature computation and model inference, an index structure is maintained to map each feature (identified by FilterID) to the physical storage addresses of its corresponding event rows.

\textbf{Feature Computation.}
When an app service (\textit{e.g.}, video recommendation) triggers an on-device model inference , the mobile device computes each required input feature in the following steps: 
(i) retrieve necessary attributes from the relevant event rows in behavior log using index structure and time window constraints, and
(ii) compute the feature value through predefined computation functions like averaging, concatenating, etc.
These resulting features are then concatenated and fed into the ML model to general final outputs for the app service.

\subsection{Storage Bottleneck of Behavior Log}
\label{sec: storage bottleneck}
Despite its critical role in supporting on-device ML, the storage overhead of behavior log has become an emerging resource bottleneck for mobile apps, driven by the increasing scale and number of ML-powered app services.
Our observation is grounded in analysis of over 20 ML models practically deployed on mobile devices, spanning service domains of live streaming, e-commerce, search and advertising domains, as detailed in~\S\ref{sec: methodology}.
\begin{figure}
    \centering
    \subfigure[Each individual model logs massive data.]{
            \includegraphics[width=0.37\linewidth]{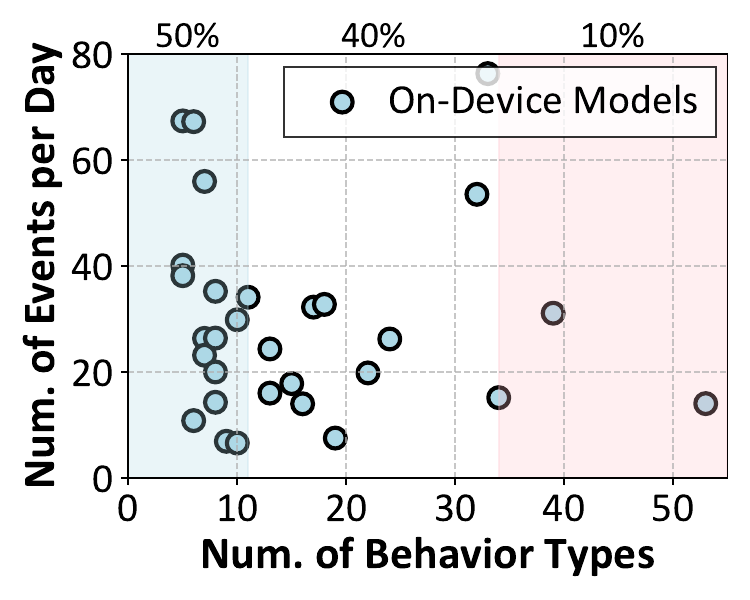}
            \label{fig: single model data}
        }
    \ \ \ 
    \subfigure[Log size grows linearly with model number.]{
        \includegraphics[width=0.37\linewidth]{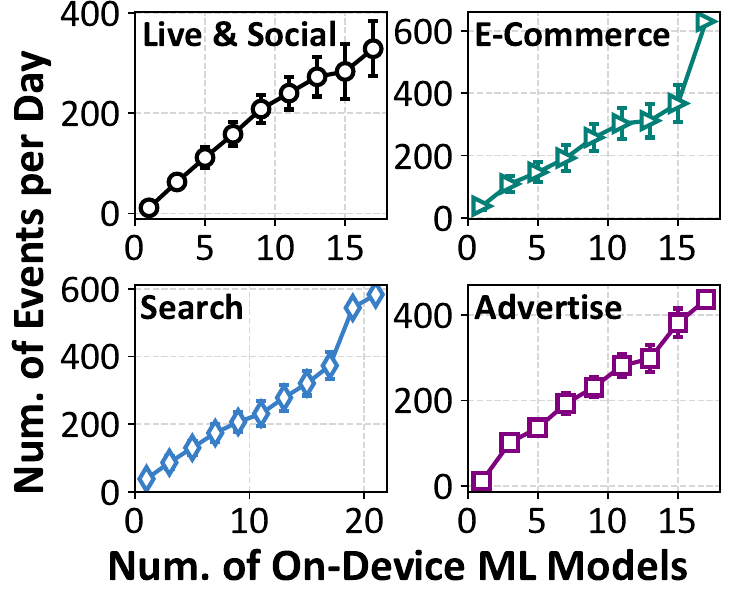}
        \label{fig: mult model data}
    }
    \caption{Statistics of real-world mobile data to illustrate behavior log storage bottleneck: (a) Each single ML model logs numerous behavior types and event rows daily; (b) Deploying more ML models linearly increases the storage cost due to their heterogeneous logging requirements.}
    \Description{Statistics of real-world mobile data to illustrate behavior log storage bottleneck: (a) Each single ML model logs numerous behavior types and event rows daily; (b) Deploying more ML models linearly increases the storage cost due to their heterogeneous logging requirements.}
\end{figure}

\textbf{Observation 1: A single model accumulates massive user behavior data.}
To comprehensively capture evolving user context and intents, each on-device ML model involves input features extracted from a wide range of user behavior types under various contexts and time windows. 
Figure \ref{fig: single model data} shows that over 50\% of the examined ML models track at least $11$ types of user behaviors and 10\% track more than 34 types. 
This leads to a daily accumulation of 5-70 (around $0.18$ MB) new event rows per model in behavior log. 
Considering that current input features can consider behaviors within time windows for up to 6 months, this accumulation significantly increases the long-term storage footprint of behavior log (around 30 MB per model).

\textbf{Observation 2: Storage overhead scales linearly with the number of ML models.}
For popular mobile apps, numerous online services are powered by ML models, such as recommendation, preloading and ranking of data with all modalities.
As each service has unique computing objective and scenario, different on-device ML models have distinct filters even for the same behavior type to compute features tailored to its task. 
For example, the video recommendation model logs only long-duration video watches to track user preferences, whereas an app exit prediction model logs all recent video watches to perceive the shifts of user attention.
Consequently, as shown in Figure \ref{fig: mult model data}, our domain-wise analysis reveals that behavior log size increases linearly with the number of on-device ML models, with each additional model contributing 20-35 new event rows per day.

\textbf{Significance of Storage Bottleneck.}
The overall storage overhead of a mobile app can be divided into two components\footnote{On iOS devices, the storage cost per app can be observed via ``Settings>General>iPhone Storage>\$App Name\$''.}: (i) \textit{App Size}, which includes the core application logic, software development kits (SDKs) and other runtime dependencies; and (ii) \textit{Document\&Data}, which stores data requiring persistent storage like chat history, cached videos and files, etc.
In practice, the app size is typically constrained to a few hundred MBs to ensure fast app launching and smooth usage experiences, while document\&data can grow to several GBs.
Unfortunately, behavior log falls under the first category as it has to be consistently kept to support latency-sensitive services at any time and cannot be arbitrarily edited by users.

As a result, behavior log can account for over $50\%$ of the app size according to our industrial data. 
The latest public statistics show that excessive app size is a primary driver of app uninstallation~\cite{app_uninstall} and leads to significant financial losses for service providers~\cite{numminen2022impact, roma2016revenue}. 
The reasons are many-facet. First, when the total size of an app becomes excessive, mobile OS platforms often issue warnings, prompting users to manually uninstall large apps to free up device storage. 
Also, a larger app size caused by behavior log can directly contribute to longer app launch time and degrade user experience.
Further, our industrial data reveals that for every additional 10 MB in app size, the number of daily active users decreases by around $30,000$ for Douyin and $61,000$ for TikTok, leading to over \$7,000 financial loss per day. 
This highlights a fundamental dilemma: while on-device ML models enables personalized and responsive user experiences, it simultaneously imposes growing storage burdens that can degrade user retention. 
By effectively relieving the pervasive mobile storage bottleneck, app developers could adopt more sophisticated and numerous ML-powered mobile services, which translates into a perceptibly superior user experience and thus drives user retention and recruitment.
Consequently, optimizing the storage of behavior log has become a critical obstacle for the broader adoption of on-device ML.

\subsection{Optimization Opportunities and Design Choices}
\label{sec: optimization opportunities}
Our investigation into behavior logs of mobile apps reveal two critical characteristics, \textit{feature-level correlation} and \textit{behavior-level heterogeneity}, which expose significant opportunities for improving storage efficiency.
These insights stem from our observations on large-scale industrial deployments and empirical analysis of real-world mobile data, in collaboration with our enterprise partner. We uncover substantial redundancy and sparsity in current behavior log structures that can be effectively minimized without information loss.
\begin{figure}
    \centering
    \subfigure[Redundant event rows.]{
        \includegraphics[width=0.35\linewidth]{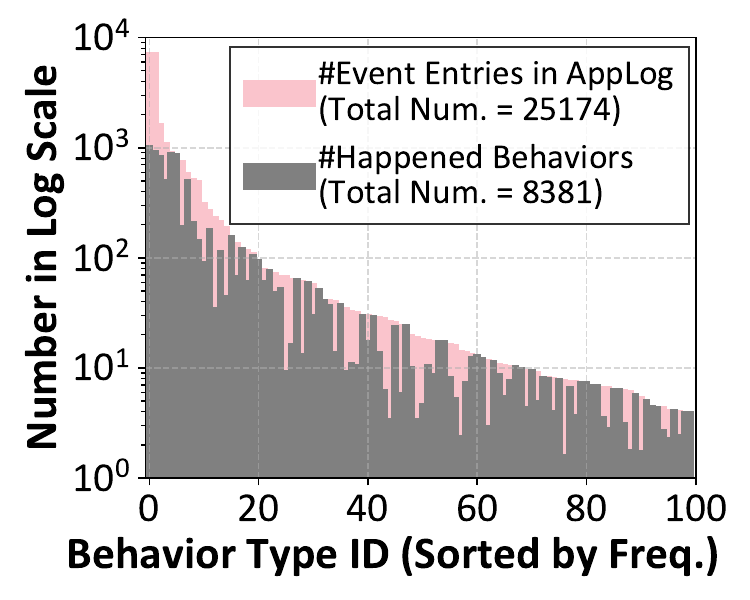}
        \label{fig: storage redundancy}
    }
    \ \ \ 
    \subfigure[Sparse attribute columns.]{
        \includegraphics[width=0.35\linewidth]{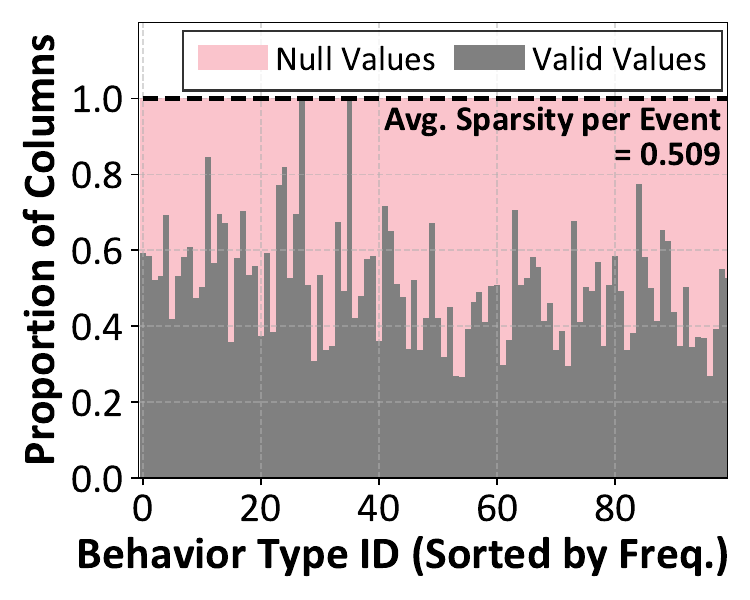}
        \label{fig: storage sparsity}
    }
    \caption{Storage redundancy and sparsity of behavior log for an average user, where x-axis denotes different behavior types and y-axis depicts their redundancy and sparsity.}
    \Description{Storage redundancy and sparsity of behavior log for an average user, where x-axis denotes different behavior types and y-axis depicts their redundancy and sparsity.}
\end{figure}

\textbf{Feature-Level Correlation.}
While input features apply distinct filters to extract respective necessary attributes from relevant behavior events, we find that their filtering conditions often overlap rather than being completely disjoint. This results in the same behavior data being recorded multiple times by different features, each associated with an individual event row and a unique FilterID.
To quantify this redundancy, we analyze behavior logs collected from a production-scale video app, Douyin(TikTok), which includes around 250 different types of user behaviors\footnote{Details on the top 50 behavior types are presented in Appendix for justification~\ref{appendix user behavior}.}.
As shown in Figure~\ref{fig: storage redundancy}, the behavior log accumulates an average of 25,174 event rows over 14 days, yet only 8,381 unique behavior events occurred in that period, implying up to 67\% redundant data storage. This highlights a substantial optimization opportunity that can be achieved by eliminating redundant data across features. 

	To reduce such data redundancy without information loss, we have two distinct design choices: value-level optimization and row-level optimization. 
The first approach maps identical attribute values into smaller symbols, which reduces the size of individual attribute value but requires maintaining an additional mapping dictionary.
The second approach merges redundant rows corresponding to the same behavior event into a unified row, reducing the total count of attribute values in behavior log. Our work considers row-level optimization due to its higher optimization potential. Value-level optimization is effective when the stored data exhibits high value-wise redundancy, \textit{i.e.}, massive attributes of different behaviors events have identical and large-size values. However, in practice, different user behaviors have quite heterogeneous attributes with diverse types and values, and each value has only small size.

\textbf{Behavior-Level Heterogeneity.}
Different types of user behaviors are inherently heterogeneous and have distinct sets of attributes. However, in current industrial practices, all behavior data is stored together in a single unified log file for index simplicity and centralized data access. 
This one-size-fits-all format introduces high storage sparsity, as irrelevant attributes are represented with null values. Our analysis of the 14-day dataset shows that: 
(i) Over 95\% of event rows contain at least one null-valued attribute,
(ii) On average, 50.9\% of all attribute values per event row are null.
While each null occupies only 1 B of space, massive event rows and numerous attributes result in a 11\% wasted storage per user.

To reduce storage sparsity, we have three design choices: sparse data encoding, column-level and row-level splitting.   
Sparse data encoding aims to reduce the physical space occupied by each null value using special markers with smaller sizes. 
Column-level splitting involves splitting the log file based on column similarity, \textit{i.e.}, grouping attributes according to their null value distributions across behavior events and storing each attribute group in one log file.
Similarly, row-level splitting decomposes the log file into multiple sub-logs based on row similarity, \textit{i.e.}, event rows that share a similar set of non-null attributes are stored densely in one log file. 
The first approach is ineffective in our context, as null values in mobile databases are already represented efficiently and the sparse storage cost is mainly caused by numerous null values rather than their individual size. 
The second approach also fails as diverse user behaviors with heterogeneous attributes make their null distributions differ significantly.
As a result, our work considers row-level splitting due to its optimal performance with the help of our dedicated designs.
\section{AdaLog Design}
We introduce {\tt AdaLog}, a system designed to optimize the storage efficiency of behavior log for ML-embedded mobile apps by reducing storage redundancy and sparsity while remaining compatible with existing on-device ML pipelines. In this section, we first provide an overview of {\tt AdaLog}~(\S\ref{sec: overview}) and then elaborate its each key component, including feature-level data merging (\S\ref{sec: feature-level data merging}), behavior-level log splitting (\S\ref{sec: behavior-level log splitting}) and incremental update mechanism (\S\ref{sec: behavior log reconstruction at scale}).

\subsection{Overview}
\label{sec: overview}
\begin{figure}
    \centering
    \includegraphics[width=0.7\linewidth]{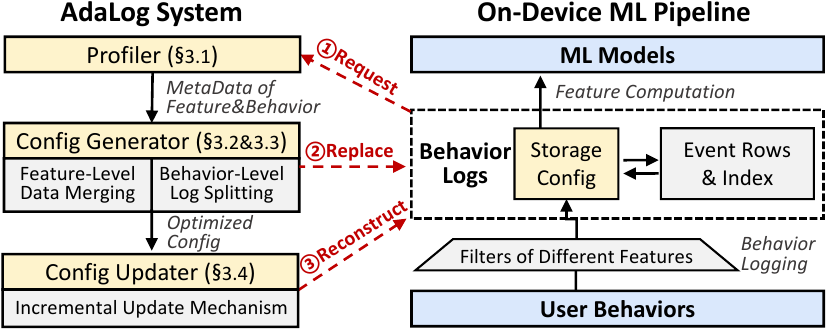}
    \caption{{\tt AdaLog} architecture and workflow within on-device ML pipelines.}
    \Description{{\tt AdaLog} architecture and workflow within on-device ML pipelines.}
    \label{fig: overview}
\end{figure}

As depicted in Figure \ref{fig: overview}, {\tt AdaLog} works as a shim layer atop existing on-device ML pipelines, making it a flexible and generalized solution without requiring modifications to device operating systems or mobile inference engines. 

\textbf{Architecture.}
As illustrated in the left part of Figure \ref{fig: overview}, {\tt AdaLog} consists of three main components that work together to optimize behavior log storage.

\noindent $\bullet$ \textit{Profiler}: 
It monitors and collects two types of lightweight metadata from behavior log, including 
(i) IDs of event rows that are logged for each feature's filter (i.e., FilterID), and 
(ii) each behavior type's attributes that have to be logged for different filters.
This metadata enables {\tt AdaLog} to analyze both feature-level data redundancy and behavior-level attribute heterogeneity, providing foundation for subsequent optimization.

\noindent $\bullet$ \textit{Config Generator}:
Given the profiler's metadata, it computes an optimal storage configuration, consisting of two aspects: 
(i) Feature-level data merging: identify which features should have their redundant event rows merged to reduce redundancy with minimal extra cost; 
(ii) Behavior-level log splitting: determine which log file to store each behavior type's event rows to eliminate sparsity. \\
$\bullet$ \textit{Config Updater}:
It applies the new storage configuration to existing behavior log in a resource-efficient manner by performing incremental updates. Through matching the current and previous storage configurations, it reuses as much existing data as possible to minimize I/O operations.

\textbf{Workflow.} As shown in the right part of Figure \ref{fig: overview}, {\tt AdaLog}'s workflow is integrated into existing on-device ML pipeline through the following two stages, ensuring compatibility and efficiency.\\
$\bullet$ \textit{Data Processing for Existing Pipeline}: 
During behavior logging, each behavior event is filtered as usual by different features. Instead of storing separate event rows for each filter, {\tt AdaLog} merges multiple features' attribute subsets into a single event row according to the feature-level data merging configuration, and stores the merged event rows in log files designated by the behavior-level splitting configuration.
When computing features for model inference, {\tt AdaLog} uses the current storage configuration to retrieve necessary attributes of relevant event rows from the appropriate log file, ensuring correctness and efficiency.\\
$\bullet$ \textit{Periodic Behavior Log Optimization}:
{\tt AdaLog} periodically (e.g., daily or during app updates allowed by users) invokes a lightweight behavior log reconstruction process, involving:
\ding{172} requesting the profiler to update metadata based on the latest behavior log,
\ding{173} replacing the outdated configuration with the new one computed by config generator,
\ding{174} incrementally updating the behavior log to adapt to the new configuration.

\subsection{Feature-Level Data Merging: Reduce Redundant Event Rows}
\label{sec: feature-level data merging}
\begin{figure}
    \centering
    \subfigure[Event rows and index structure before feature-level data merging.]{
        \includegraphics[width=0.7\linewidth]{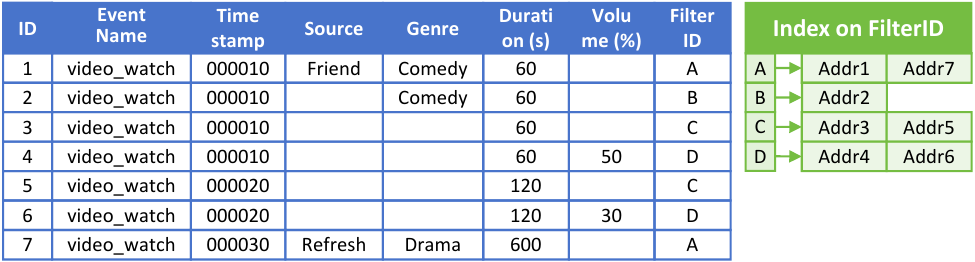}
        \label{fig: data and index storage}
    }
    \subfigure[Event rows and index structure after merging all redundant event rows, where: 7-3=4 repetitive rows are removed, while 12-7=5 additional addresses are recorded by index.]{
        \includegraphics[width=0.75\linewidth]{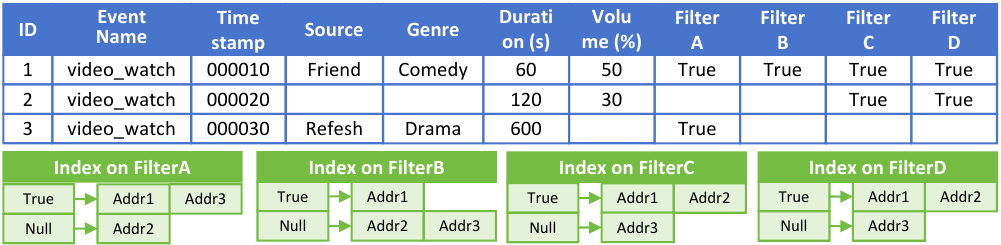}
        \label{fig: behavior log after data merging}
    }
    \caption{Comparing behavior logs before and after conducting feature-level data merging.}
    \Description{Comparing behavior logs before and after conducting feature-level data merging.}
    \label{fig: example for data merging}
\end{figure}
A straightforward method to eliminate redundancy in behavior log is to merge all event rows across features that correspond to the same behavior event. As illustrated in Figure \ref{fig: example for data merging}, this involves two steps:
(i) Merging event rows with identical event name and timestamp attributes into a single row containing the union of all required attributes;
(ii) Appending a set of FilterID columns to the merged row to differentiate which filters the merged row satisfies.
While this method effectively eliminates redundant data, it introduces a new significant challenge: index inflation.

In modern mobile apps, behavior logs are indexed on FilterID attribute column to support fast data retrieval for computing each feature. 
As shown in the green part of Figure \ref{fig: example for data merging}, the database index structure maps every value of the indexed column, including nulls, to the physical addresses of the rows where those values appear. 
However, since each behavior event satisfies only a subset of features' filters, the merged rows inevitably contain null values in the appended FilterID columns~(shown in blue part of Figure \ref{fig: behavior log after data merging}). These nulls inflate the index structure by forcing it to store unnecessary mappings~(shown in green part of \ref{fig: behavior log after data merging}).
Consequently, if redundant data is merged across inappropriate features, the overhead of the expanded index can easily surpass the storage saved by eliminating redundant data. This trade-off becomes more severe with massive models and features, as the number of possible feature groups grows exponentially. 

To address this issue, we formulate the feature merging decision as a classic maximum weighted matching problem in a hypergraph, and design a hierarchical merging algorithm with polynomial time complexity for scalable on-device execution.
\begin{figure}
    \centering
    \includegraphics[width=\linewidth]{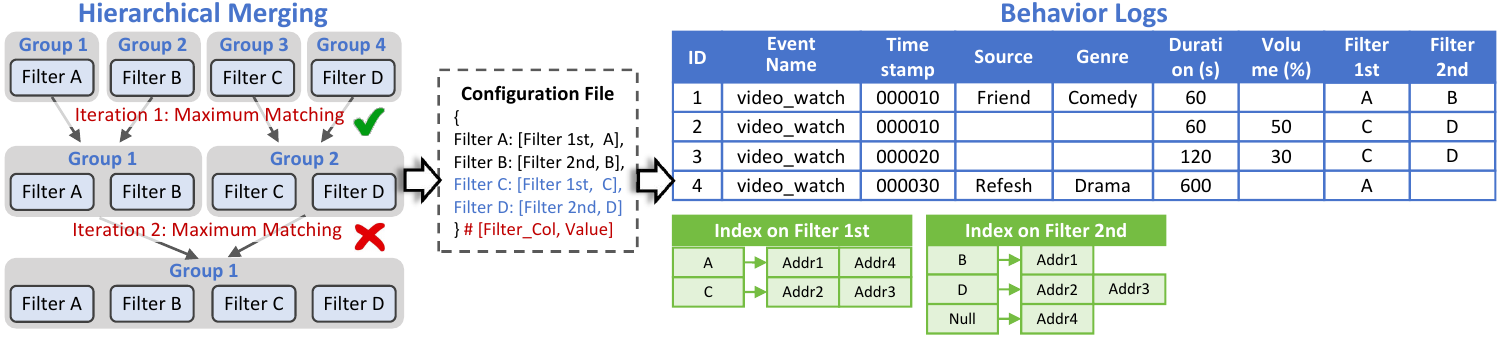}
    \caption{An example to illustrate hierarchical merging algorithm (left), storage configuration of feature merging (middle) and behavior log after merging (right). Compared to Figure \ref{fig: data and index storage}, 7-4=3 redundant event rows are eliminated while only 8-7=1 address is additionally recorded by index.}
    \Description{An example to illustrate hierarchical merging algorithm (left), storage configuration of feature merging (middle) and behavior log after merging (right). Compared to Figure \ref{fig: data and index storage}, 7-4=3 redundant event rows are eliminated while only 8-7=1 address is additionally recorded by index.}
    \label{fig: hierarchical merging illustration}
\end{figure}

\textbf{Problem Formulation.}
Given a set of features $\mathcal{F}$ and each feature $f\!\in\!\mathcal{F}$'s relevant event rows $\mathrm{E}(f)$, attributes $\mathrm{A}(f)$ and physical address size $\mathrm{Size}(Addr)$, we aim to minimize the total storage cost by partitioning features into disjoint feature groups $\mathcal{G}\!=\!\{g_1,\dots,g_M\}$ where intra-group features share merged event rows. The optimization problem can be formally expressed as:
\begin{equation}
        \begin{aligned}
            & \mathcal{G}^*=\mathop{\arg\min}_{\mathcal{G}:\ \cup_{g\in \mathcal{G}}g=\mathcal{F}} \sum_{i=1}^M \big[\mathrm{Data\_Size}(g_i)+\mathrm{Index\_Size}(g_i)\big],\\
            \mathrm{s.t.}\ & \mathrm{Data\_Size}(g_i)=\underbrace{\big|\cup_{f\in g_i}\mathrm{E}(f)\big|}_\text{Num. of Event Rows}\times \underbrace{\mathrm{Size}\big(\cup_{f\in g_i}\mathrm{A}(f)\big)}_\text{Size per Event Row},\\
            & \mathrm{Index\_Size}(g_i)=\underbrace{|\cup_{f\in g_i}E(f)|\times\mathrm{Size}(Addr)}_\text{Address Size per Index}\times \underbrace{\max_{g\in \mathcal{G}}|g|}_\text{Index Num}.
        \end{aligned}
    \nonumber
    \label{eq: data merging problem formulation}
\end{equation}
For each feature group $g_i$, data\_size captures the space for storing event data, measured as the product of the number of event rows and the size per row, while index\_size represents the index structure space, quantified as the product of the number of indexed columns and the total address size of all event rows. 

\textbf{NP-Hardness.}
We notice that this problem can be interpreted as a maximum weighted matching problem in a hypergraph $G\!=\!(V, E)$, where:
(i) Each feature $f\!\in\!\mathcal{F}$ is represented as as a node $v\!\in\!V$;
(ii) Each potential feature group $g\!\subseteq\!\mathcal{F}$ is represented as a hyperedge $e\!\in\!E$ connecting its member features' nodes;
(iii) The weight of hyperedge $e$ equals the overall storage savings from merging features in $g$.
A valid feature grouping strategy corresponds to a matching in the hypergraph, \textit{i.e.}, a set of disjoint hyperedges. 
Thus, finding the optimal feature grouping strategy is equivalent to solving the maximum weighted matching problem on the hypergraph $G$, a well-known NP-hard problem in graph theory~\cite{DBLP:conf/soda/CyganGM13, lozin2008polynomial, DBLP:journals/dam/BrandstadtM18a}.

\textbf{Hierarchical Merging Algorithm.}
To efficiently solve the NP-hard hypergraph-based problem in practical mobile settings, we propose a hierarchical merging algorithm that avoids direct hypergraph optimization by decomposing it into a series of tractable 2D-graph matchings.
This is because maximum matchings in 2D-graphs can be found by Blossom algorithm~\cite{kolmogorov2009blossom} with only polynomial $O(|V|^3)$ complexity.
Our key idea is similar to hierarchical clustering~\cite{murtagh2012algorithms} as shown in the left part of Figure \ref{fig: hierarchical merging illustration}: we start from a fine-grained configuration where each feature is treated as a standalone group and iteratively merge \textit{pairs of feature groups} that lead to the highest storage reduction, providing a practical balance between optimization quality and system efficiency. 

Specifically, in each iteration $t$, we construct a weighted 2D-graph $G=(V,E)$ for current feature groups $\mathcal{G}^t$, where each node $v\in V$ denotes a feature group $g\in\mathcal{G}^t$, each edge $e=(g_i,g_j)\in E$ connects two feature groups $g_i$ and $g_j$ and the edge weight $w(g_i,g_j)$ quantifies the storage savings if event rows belonging to feature groups $g_i$ and $g_j$ are merged:
\begin{equation}
        \begin{aligned}
            & w(g_i',g_j') = \Delta \text{Data\_Size} + \Delta\text{Index\_Size}\approx\\
            & \underbrace{|\text{E}(g_i')\!\cap\!\text{E}(g_j')|}_\text{Redundant Rows} \times 
            \underbrace{\Big[\text{Size}\big(A(g_i')\big)\!+\!\text{Size}\big(A(g_j')\big)\!-\!\text{Size}\big(A(g_i')\!\cup\! A(g_j')\big)\Big]}_\text{Size of Overlapped Attributes per Row}\\
            & + \Big[\underbrace{|\mathrm{E}(g_i')\!\cup\!\mathrm{E}(g_j')|\times |g_i'\!\cup\! g_j'|}_\text{Address Num. after Merging} - \underbrace{\big(|\mathrm{E}(g_i')|\times|g_i'| + |\mathrm{E}(g_j')|\times|g_j'|\big)}_\text{Address Num. Before Merging} \Big] \times \mathrm{Size}(Addr).
        \end{aligned}
    \nonumber
\end{equation}
The data\_size term captures data storage reduction due to eliminating attributes of overlapping event rows across features and the index\_size term accounts for index size changes.
Then, using the Blossom algorithm, we identify a maximum weighted matching on the 2D-graph $G$, \textit{i.e.}, a set of disjoint pairs $(g_i, g_j)$ with the highest storage savings. If the total gain is positive, the matched pairs are merged into new feature groups $\mathcal{G}^{t+1}$ for next iteration. Otherwise, the algorithm terminates and outputs current feature groups as feature-level data merging configuration.

The hierarchical merging algorithm runs for at most $\log_2|\mathcal{F}|$ iterations, as the number of feature groups (\textit{i.e.}, nodes in the graph) is halved in each iteration, implying $\frac{|\mathcal{F}|}{2^{t-1}}$ nodes in the $t$-th iteration. Consequently, the total time complexity becomes polynomial:
\begin{equation}
        \sum_{t=1}^{\log_2|\mathcal{F}|}O\left(\left(\frac{|\mathcal{F}|}{2^{t-1}}\right)^3\right)\!=\!O(|\mathcal{F}|^3)\!\cdot\!\sum_{t=1}^{\log_2|\mathcal{F}|}\left(\frac{1}{8}\right)^{t-1}\!=\!O(|\mathcal{F}|^3).
    \nonumber
\end{equation}

\textbf{System Implementation and Optimization.}
The hierarchical merging algorithm can be integrated into {\tt AdaLog} with further system optimization. First, {\tt AdaLog}'s profiler can directly collect IDs of event rows relevant to each feature using the index structure. Other necessary information such as attribute size and address size is fixed and can be profiled in advance. 
Second, {\tt AdaLog}'s generator computes the optimal merging configuration in an efficient manner. It pre-clusters features based on the targeted behavior types and performs hierarchical merging algorithm independently for each behavior type's related features, which reduces the problem size and enables parallel execution.
As shown in Figure \ref{fig: hierarchical merging illustration}, the configuration is stored in a dictionary-like structure, which designates (i) features in the same group to distinct FilterID columns and (ii) features across different groups to shared FilterID columns with different specific values.
Our evaluations in \S\ref{sec: overall performance} demonstrate that {\tt AdaLog} incurs $\le\!1$ second of latency to complete the entire algorithm on device and the configuration size is $\le\!10$ KB, demonstrating high system efficiency.

\subsection{Behavior-Level Log Splitting: Minimize Overall Sparsity}
\label{sec: behavior-level log splitting}
\begin{figure}
    \centering
    \subfigure[Heterogeneous attribute distribution.]{
        \includegraphics[height=3.5cm]{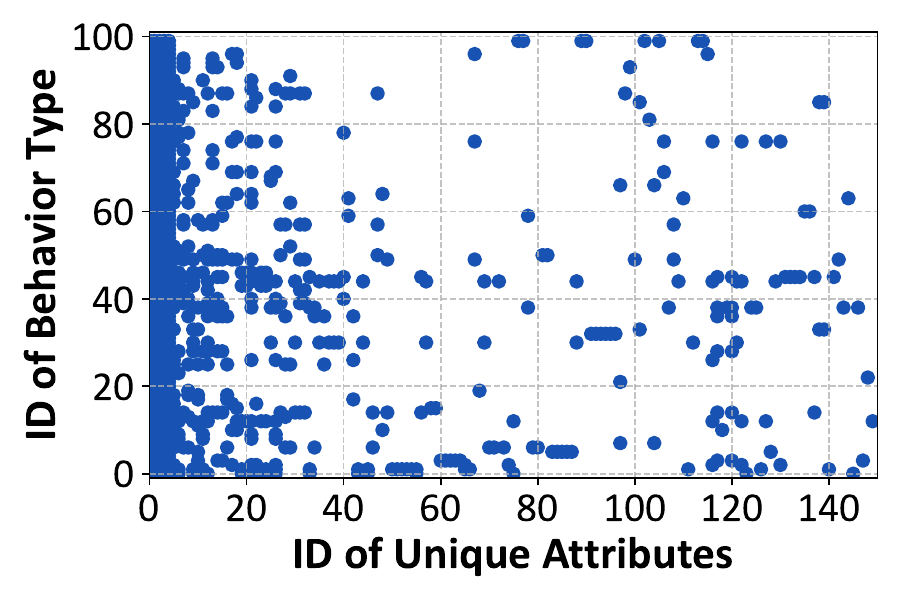}
        \label{fig: event attribute distribution}
    }
    \ \ \ 
    \subfigure[Similar attribute number.]{
        \includegraphics[height=3.5cm]{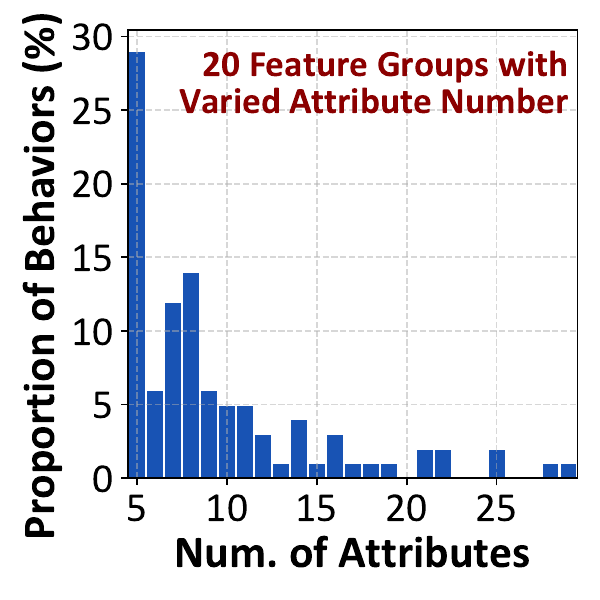}
        \label{fig: event attribute group}
    }
    \caption{Different user behavior types have heterogeneous attributes but similar attribute number.}
    \Description{Different user behavior types have heterogeneous attributes but similar attribute number.}
\end{figure}

In modern mobile apps, hundreds types of user behaviors are captured as heterogeneous events, each containing a unique set of behavior-specific attributes, as shown in Figure \ref{fig: event attribute distribution}. 
Current industrial practices of app behavior logs commonly store all behavior events in a single unified log file to simplify indexing, querying and management. However, this approach results in severe storage sparsity when more types of user behaviors are consumed by ever-growing ML-embedded services, as each event row can contain massive null values for irrelevant attributes.

A direct solution to eliminate sparsity is behavior-level log splitting, where behavior events are stored in separate log files according to their attribute sets, ensuring that all event rows within one log file share identical attributes. However, this solution is impractical: the heterogeneous attribute sets of different behavior types require splitting the behavior log into hundreds of small, fragmented files, as illustrated in Figure \ref{fig: event attribute distribution}.
While this strategy minimizes storage sparsity, it introduces significant overhead for file management and metadata storage, as each small file requires its own set of metadata such as table names, column names, index structures, file sizes, etc. 

\textbf{Virtually Hashed Attribute Name}.
To overcome the above challenge, we propose virtually hashed attribute naming (VHAN) design, a logically sparse but physically dense storage design for behavior logs to reduce sparsity without creating massive files.
We observe that different behavior types often have similar numbers of attributes but different attribute names (Figure \ref{fig: event attribute group}), which is the root cause preventing them from storing in the same dense log file.
Therefore, we propose to decouple the storage of attribute values from their physical names by using virtual attribute IDs. This design is analogous to virtual memory in operation systems~\cite{denning1970virtual} where virtual addresses abstract away physical memory locations. 
\begin{figure}
    \centering
    \includegraphics[width=0.65\linewidth]{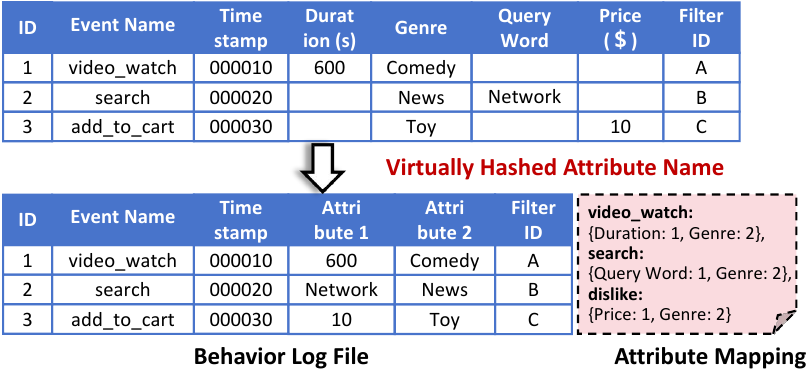}
    \caption{VHAN uses attribute mapping to densely storage behavior events with identical attribute numbers.}
    \Description{VHAN design uses attribute mapping to densely storage behavior events with identical attribute count.}
    \label{fig: virtually hashed attribute name}
\end{figure}
As shown in Figure \ref{fig: virtually hashed attribute name}, VHAN consists of two main parts: 
\begin{itemize}[topsep=0cm, leftmargin=0.35cm]
    \item \textit{Attribute Mapping}: A dictionary-like configuration file that maps each physical attribute name into a virtual ID for each user behavior type.
    \item \textit{Behavior Log File}: The physical storage for event rows, which retains the same structure as existing behavior log but replaces attribute names with their corresponding virtual IDs from the attribute mapping.
\end{itemize}
By leveraging VHAN, {\tt AdaLog} enables the dense storage of any behavior events with identical numbers of attributes in one behavior log file, eliminating the strict requirement of totally same attribute sets.
As a result, we propose to cluster user behaviors according to their cardinality of attribute sets, and store event rows of behaviors within the same cluster in one log file, which reduces the number of log files from the number of unique attribute sets ($\approx$250) to the number of unique attribute count ($\approx$20), as shown in Figure \ref{fig: event attribute group}.

\textbf{System Implementation Overhead.}
The VHAN design introduces an additional attribute mapping file, which is used during behavior logging and feature computation to map physical attribute names to virtual IDs for event storage and attribute retrieval.
Therefore, two choices of overhead are introduced:
\textit{(i) Memory}: Maintaining the attribute mapping in device memory facilitates real-time event logging and feature computation, but introduces a memory footprint of around $30$KB, which is negligible for even low-end smartphones.
\textit{(ii) Latency}: Alternatively, loading the attribute mapping on-demand for each event logging or model inference process introduces millisecond-level latency, which is typically acceptable for most real-time applications.

\subsection{Behavior Log Reconstruction at Scale}
\label{sec: behavior log reconstruction at scale}
The previous designs successfully optimize the storage efficiency for given static behavior data. However, mobile users' behavior patterns are inherently dynamic and unpredictable, which presents significant scalability challenges in maintaining up-to-date configurations and behavior logs in real-world mobile settings.

Specifically, as new user behavior events are continuously recorded in behavior log, the distribution of event rows across features also shifts. This leads to changes in the optimal storage configurations and requires reconstruction of the behavior log.
However, reconstruction typically incurs substantial latency due to the large volume of event rows in behavior logs. For example, reconstructing a 10 MB behavior log on iPhone 13Pro requires around 10 seconds. This can lead to intolerant app performance degradation due to preempting computation and I/O resources necessitated by other concurrent services, like video rendering and content loading.
\begin{figure}
        \subfigure[Reconstruction time breakdown.]{
            \includegraphics[width=0.31\linewidth]{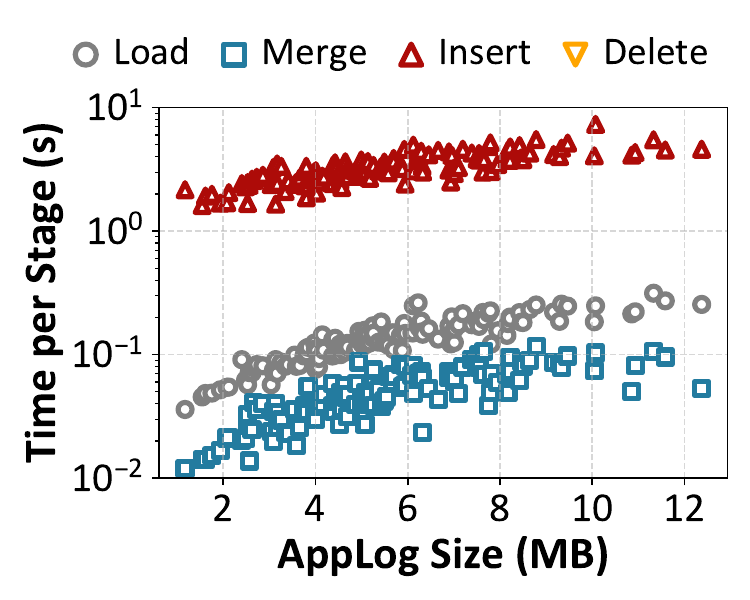}
            \label{fig: update time breakdown}
        }
        \ 
        \subfigure[Accumulated config. changes.]{
            \includegraphics[width=0.31\linewidth]{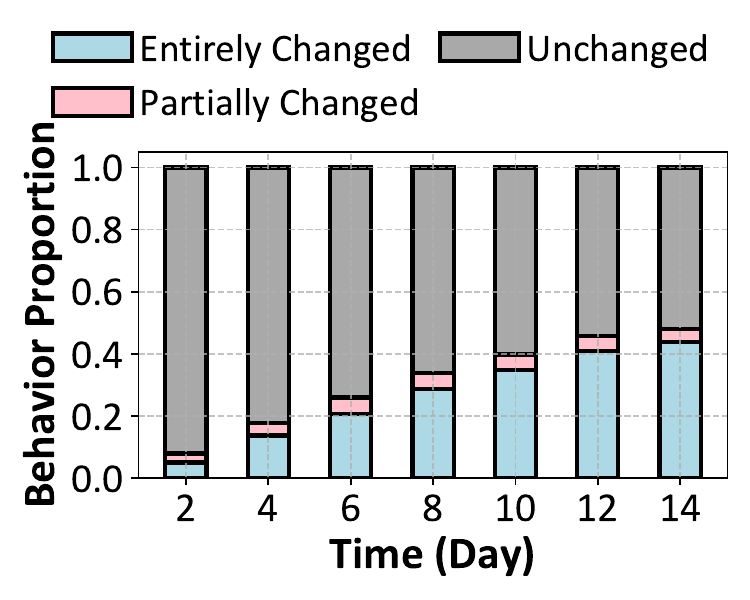}
            \label{fig: accumulate configuration}
        }
        \ 
        \subfigure[Incremental config. changes.]{
            \includegraphics[width=0.31\linewidth]{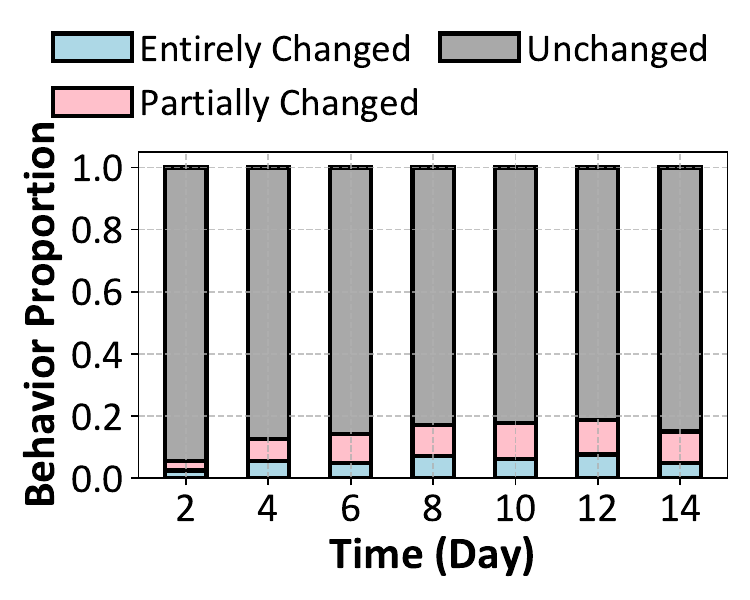}
            \label{fig: incremental configuration}
        }
        \caption{Analysis of real-world behavior logs to (a) break down reconstruction time, and (b)(c) reveal the proportions of an average user's behaviors that experience changes in storage configuration over time.}
        \Description{Analysis of real-world behavior logs to (a) break down reconstruction time cost, and (b)(c) reveal the proportions of an average user's behaviors that experience configurations changes over time.}
\end{figure}

To address this challenge, we start with analyzing the bottleneck operations during the log reconstruction process, and further propose an incremental update mechanism that allows {\tt AdaLog} to adapt existing behavior log files to new configurations with minimal system overhead.

\textbf{Overhead Breakdown.}
The reconstruction of behavior logs involves four major steps: 
\textit{(i) Loading} relevant event rows for each feature, 
\textit{(ii) Merging} these rows according to the feature-level data merging strategy, 
\textit{(iii) Inserting} the merged rows into newly created behavior log files determined by the behavior-level log splitting configuration,
\textit{(iv) Deleting} outdated log files, which incurs negligible cost.
To analyze the bottleneck operations, we conducted an extensive analysis of hundreds of real-world behavior log reconstruction workloads across various file sizes. As shown in Figure \ref{fig: update time breakdown}, we find that approximately 95\% of the reconstruction time is dominated by I/O-intensive operations, such as data loading and inserting. This observation motivates us to minimize unnecessary I/O operations by reusing as much of the existing data as possible, thereby enabling incremental updates instead of full reconstructions.

\textbf{Incremental Update Opportunities.}
To assess the feasibility and potential of incremental updates, we investigate how the optimal storage configuration of a mobile user evolves over time. Specifically, we collect daily optimal configurations for TikTok users over a 14-day period and decompose them into behavior-wise configurations. Each behavior's configuration is then categorized into three types:
\begin{itemize}[topsep=0cm, leftmargin=0.35cm]
    \item \textit{Entirely Changed}: Both data merging and log splitting configurations are altered, requiring a full reconstruction of corresponding event rows.
    \item \textit{Partially Changed}: Only specific feature groups within the data merging configuration are modified, requiring updates to the affected event rows.
    \item \textit{Unchanged}: The configuration remains stable, and no updates are needed.
\end{itemize}
As shown in Figure \ref{fig: accumulate configuration}, while the proportion of behaviors with entirely changed configurations accumulates over time, most behaviors' configurations are unchanged or partially modified. When examining changes over shorter intervals, such as a 2-day window (Figure~\ref{fig: incremental configuration}), 86\% of behaviors retain the same configuration, 10\% exhibit partial changes, and only 4\% require a complete reconstruction.
These findings suggest that most configuration changes can be handled through incremental updates rather than full reconstructions, leading to significant optimization potential.

\textbf{Incremental Update Mechanism.}
Building on the above analysis of configuration evolution, we propose a novel incremental update mechanism for {\tt AdaLog}'s updater module. Our core idea is to reuse as many of the existing event rows as possible and minimize the I/O operations required to adapt them to new configurations.
As elaborated in Figure \ref{fig: updater design}, our incremental updater operates through three key steps:
\begin{figure}
    \begin{minipage}[b]{0.6\linewidth}
        \includegraphics[width=\linewidth]{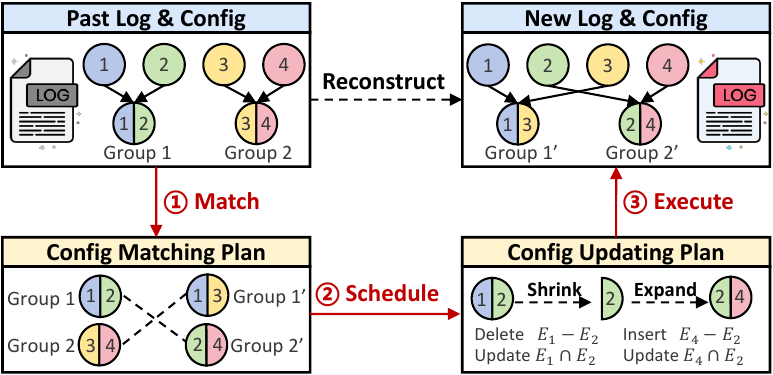}
        \caption{{\tt AdaLog}'s Incremental update mechanism.}
        \Description{{\tt AdaLog}'s Incremental update mechanism.}
        \label{fig: updater design}
    \end{minipage}
\end{figure}

\textit{\ding{172} Match.}
To efficiently adapt the behavior log with the old configuration to the new one, we model the optimal adaption process as a maximum weighted matching problem in a bipartite graph:
\begin{itemize}[topsep=0cm, leftmargin=0.5cm]
    \item Each node in the left partition of the bipartite graph represents a feature group $g_i \in \mathcal{G}$ in the old configuration.
    \item Each node in the right partition represents a feature group $g_j' \in \mathcal{G}'$ in the new configuration.
    \item Each pair of feature groups $(g_i, g_j')$ is connected by an edge whose weight is defined as the cardinality of the intersection between their event rows, i.e., $|E(g_i) \cap E(g_j')|$.
\end{itemize}
The intuition behind this formulation is that a larger data overlap between two feature groups implies greater potential for data reuse. By maximizing the total weight of matched pairs, we minimize the number of I/O operations required for insertion, deletion, or attribute updates, since a high-overlap pair can be transformed with minimal modification.
Next, we apply the Blossom algorithm to determine the optimal one-to-one mapping between old and new feature groups\footnote{For each feature group in the new configuration: If it is in the matching, we transform event rows of the matched past feature group; Otherwise, we simply reconstruct the required event rows.}, ensuring that the transformation plan preserves as much existing data as possible.

\textit{\ding{173} Schedule.} 
Once the optimal mapping between old and new feature groups is obtained, we introduce a shrink-and-expand strategy to transform event rows from the old configuration to the new one with minimal rewriting overhead:
\begin{itemize}
    \item \underline{(i) Shrink}: For each matched pair $(g_i, g_j')$, we reduce $g_i$ to the intersection $g_{ij'} = g_i \cap g_j'$. This involves removing event rows from $E(g_i)$ that are not present in $E(g_{ij'})$, as well as pruning obsolete attributes from the overlapping rows $E(g_i) \cap E(g_{ij'})$.
    \item \underline{(ii) Expand}: After shrinking, we expand the pruned set $g_{ij'}$ to fully match the structure of $g_j'$. This expansion step inserts new event rows that are required by $g_j'$ but not present in $E(g_{ij'})$, and inserts newly introduced attributes to the existing rows in $E(g_{ij'}) \cap E(g_j')$.
\end{itemize}

\textit{\ding{174} Execute.}
Finally, the incremental update plan is executed for each behavior type. If the configuration has entirely changed, such that $g_{ij'} = \varnothing$, the mechanism falls back to a full reconstruction of the new feature group.
If the configuration is partially changed, we leverage the matching and scheduling plan to perform minimal updates, effectively reusing the majority of event data.

\section{Evaluation}
\label{sec: Evaluation}
In this section, we systematically evaluate the performance of {\tt AdaLog} to answer the following key questions:
How effectively and efficiently does {\tt AdaLog} reduce behavior log storage overhead across diverse mobile users and application domains (\S\ref{sec: overall performance})?
What is the contribution of each core design component to {\tt AdaLog}'s performance (\S\ref{sec: component-wise analysis})?
How is {\tt AdaLog} affected by different factors (\S\ref{sec: sensitivity analysis})?

\subsection{Methodology}
\label{sec: methodology}
\begin{figure}
    \centering
        \centering
        \subfigure[Behavior types used by different models.]{
            \includegraphics[height=3.75cm]{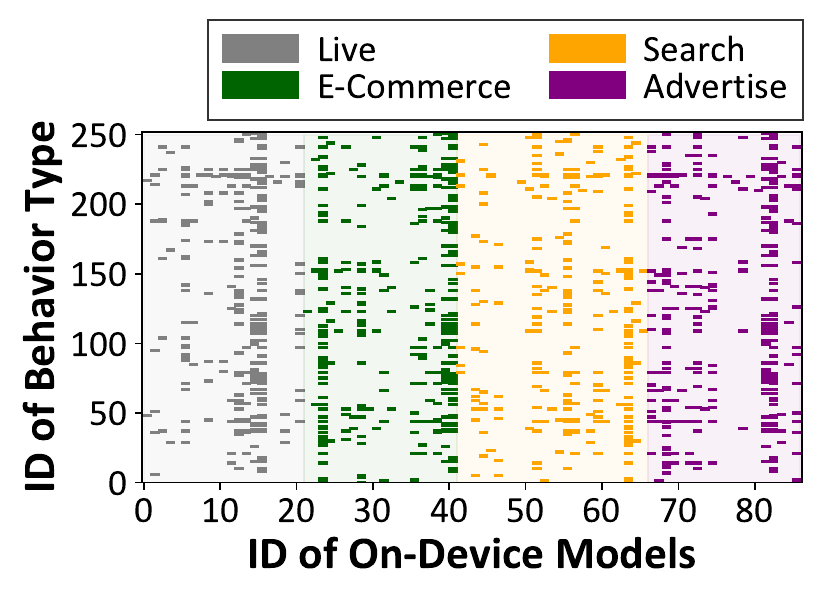}
            \label{fig: features of models}
        }
        \ 
        \subfigure[Number of event rows.]{
            \includegraphics[height=3.75cm]{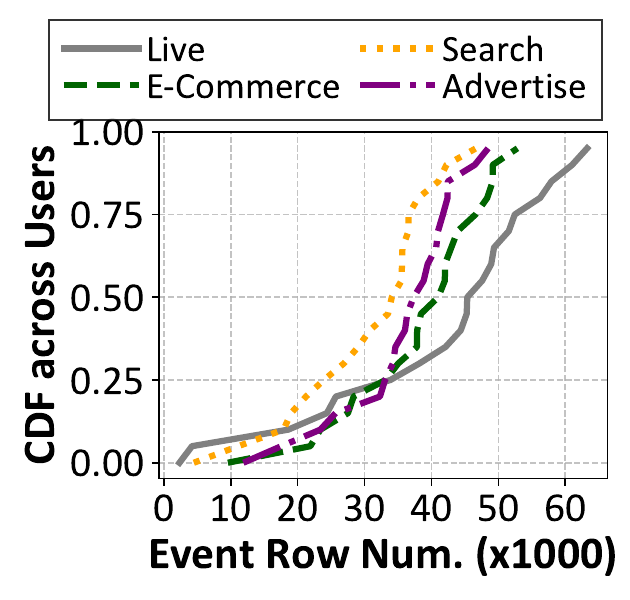}
            \label{fig: entry num distribution}
        }
        \ 
        \subfigure[Sizes of behavior log.]{
            \includegraphics[height=3.75cm]{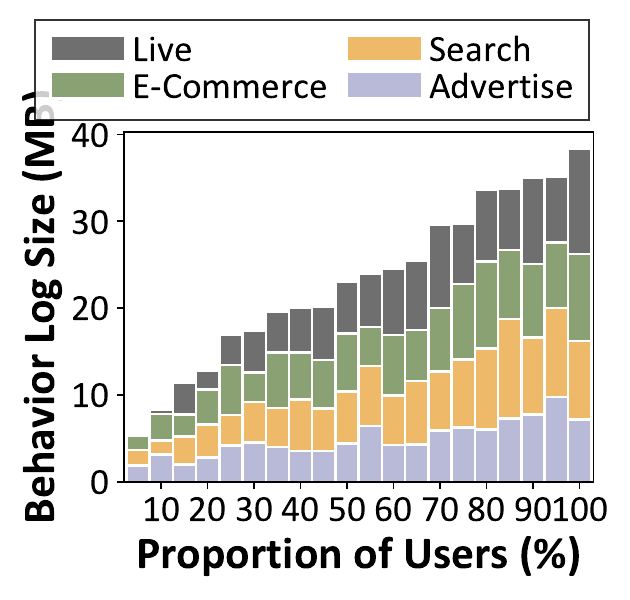}
            \label{fig: behavior log size distribution}
        }
        \ 
        \caption{High-level statistics of user behaviors, ML models and datasets involved in our evaluation. The top 50 user behaviors are presented in Appendix for justification.}
        \Description{High-level statistics of user behaviors, ML models and datasets involved in our evaluation. }
        \label{fig: high-level statistics}
\end{figure}

\textbf{Implementation.}
We have fully implemented {\tt AdaLog} as a system prototype comprising around $3K$ lines of code, 
	which was deployed and evaluated on real-world mobile apps within the ByteDance ecosystem, a company with advanced mobile AI technology and billions of daily active users.
Specifically, {\tt AdaLog} is packaged as a lightweight Python library with developer-friendly APIs 
	, and thus can be seamlessly integrated into the app SDKs of online users who consented to participate in our system evaluation and data collection.
During evaluation, {\tt AdaLog} operates autonomously to optimize behavior log on a daily basis\footnote{
	In our default evaluation setting, we consider updating each user's behavior log on a daily basis, driven by both system constraints and the temporal characteristics of user data. 
First, to avoid disrupting on-device model inference and compromising user experience, the log update process is expected to be scheduled during a guaranteed low-usage time period, typically deep at night when the mobile device is idle or charging.
Second, we observe that the impact of a single day's user interactions on the data distribution of overall behavior log is relatively limited. Updating more frequently than daily would incur more system overhead without substantial gain in log size reduction, as analyzed in \S\ref{sec: sensitivity analysis}.
} without manual intervention from users or developers.\\
\textit{It is important to note that all data collection and system operations strictly comply with privacy-preserving standards set by both industry and academia.}

\textbf{Mobile App Domains.}
To demonstrate the broad applicability of {\tt AdaLog}, we evaluate its performance across four representative mobile app domains, each involving various application services and on-device ML models. \\
\textit{$\bullet$ Live streaming} (e.g., TikTok): This domain of app involves $21$ on-device ML models for services like customized video preloading, recommendation, bandwidth management, user engagement prediction, etc. User behaviors include comments, (dis)likes, shares, subscribes, etc.\\
\textit{$\bullet$ E-Commerce} (e.g., Taobao): This domain of app employs $20$ ML models for personalized product recommendation, item ranking and preloading as well as comment selection by analyzing user-product interactions such as item clicks, favorites, adding to cart and purchases.\\
\textit{$\bullet$ Search} (e.g., Baidu): This domain of app leverages $25$ on-device ML models to improve users' searching experiences through predicting query keywords, ranking returned results, preloading multi-modal content, predicting search and exit timing, etc.\\
\textit{$\bullet$ Advertisement \& Monetization} (e.g., Google Ads): $20$ on-device ML models are used to optimize advertisement delivery, targeting and monetization for maximizing user engagement and application revenue opportunities.\\
\textit{$\bullet$ Unified Application}: An ideal case where user data across multiple app domains can be stored and optimized in an unified manner. This setup is feasible for (i) services or mobile apps belonging to the same parent company, and (ii) an operating system authorized to manage various native apps.

	Due to strict enterprise confidentiality requirements, we cannot disclose the specific names of our testing mobile apps. The number and name of ML-powered services within an mobile app are quite sensitive due to their importance to user experience guarantee and high economic profit. We acknowledge that our primary evaluation focuses on apps where the user base is predominantly Chinese, which may lead to potential biases on app usage patterns and performance evaluation.

\begin{figure}
    \centering
    \includegraphics[width=0.7\linewidth]{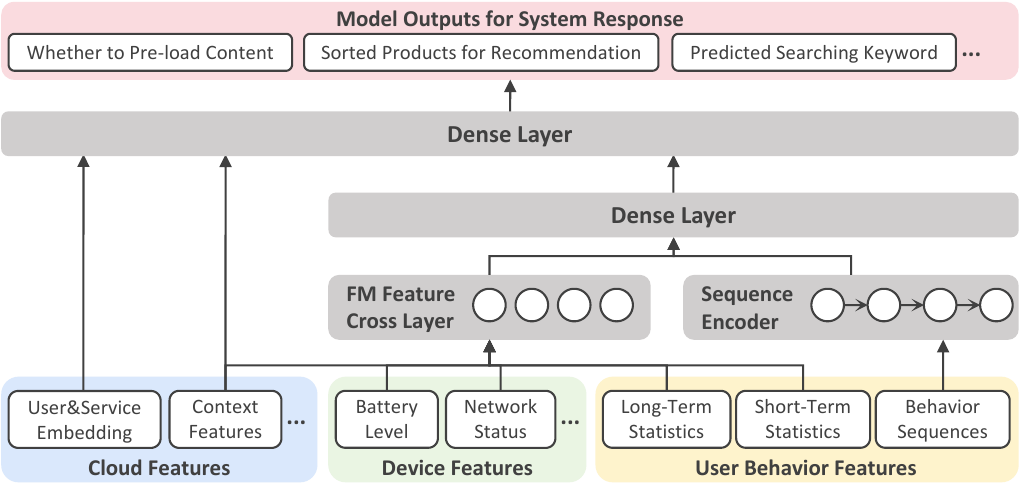}
    \caption{
    		Abstracted model architecture of common on-device machine learning models in practice.
    }
    \label{fig: model architecture}
\end{figure}

\textbf{Models.}
In our experiments, the on-device ML models span a range of complexity, from lightweight models such as decision trees~\cite{quinlan1996learning} and multilayer perceptrons (MLPs)~\cite{popescu2009multilayer} that leverage a few behavior features, to complex deep neural networks~\cite{cheng2016wide, covington2016deep, gomez2015netflix} that leverage hundreds of behavior features\footnote{
While our evaluation focuses on ML models that have been deployed within mobile apps, we believe that advanced mobile intelligence with large language models~\cite{lin2024awq, wen2024autodroid, yuan2024mobile} will require much more user behavior data for personalized finetuning and inference, making our work more applicable in the future.
}. 
We provide high-level statistics in Figure \ref{fig: features of models} to depict the user behaviors leveraged by different models and features. While we cannot disclose the specific structure of testing models, we present a general model architecture adopted by most mobile services in Figure \ref{fig: model architecture}, which composes of three layers. \textit{(i) Input Layer}: An on-device model takes three categories of features as inputs: cloud features to provide global information, device features to describe the current device state and massive behavior features to summarize various historical user behaviors; \textit{(ii) Processing Layer}: These features are then processed by different layers. Statistical features of user behaviors and device features are passed into an FM (Factorization Machine) layer for feature crossing, while sequential features of user behaviors are sent to an sequence encoder to capture temporal dynamics and periodical patterns; \textit{(iii)} Output Layer: Finally, the combined feature outputs are passed through several dense layers to generate final predictions for personalized system responses. 
The on-device ML models are typically limited to tens of MBs in size. This constraint is driven by two critical factors: \textit{(i) Model size limitation:} Mobile platforms impose strict limits on app size (e.g., 2GB for iOS~\cite{ios_space} and 4GB for Android~\cite{android_space}), which directly limits the size of each ML model deployed within a mobile app; \textit{(ii) Inference latency:} Ensuring low-latency model execution (typically within hundreds of milliseconds) is essential for a good user experience, which further limits model complexity and size. 

\textbf{Datasets.}
For each app domain, we evaluate {\tt AdaLog} using real-world behavior log collected from real-world mobile users over a 14-day period. 
The datasets include operating systems of iOS and Android, various app usage frequencies (ranging from 2,340 to 52,852 behavior events per user as shown in Figure \ref{fig: entry num distribution}) and a wide range of behavior log file sizes as shown in Figure \ref{fig: behavior log size distribution}. 
    The observed behavior log sizes are on the order of tens of MBs because the analysis is based on only 14 days of data per user, constrained by the enterprise’s online evaluation and data collection process. In real-world deployments, the log sizes would be significantly larger, as they typically record behavior events spanning much longer time periods up to 6 months.  
    Note that if 14 days of data yields substantial storage reduction, the benefits will only scale as the log size grows, as analyzed in \S\ref{sec: sensitivity analysis}.

\textbf{Baselines.}
To the best of our knowledge, {\tt AdaLog} is the first system designed to optimize behavior log storage for practical ML-embedded mobile apps. Thus, we compare {\tt AdaLog} against two baselines: 
(i) \textit{w/o AdaLog}: A standard industry behavior log system that does not incorporate {\tt AdaLog} support. In this system, all behavior events relevant to on-device ML models are stored in behavior log without optimization, leading to higher storage overhead.
(ii) \textit{AdaLog variants}: Different variants of {\tt AdaLog} where individual components are disabled or modified, allowing us to isolate the impact of each design technique on system performance.

\textbf{Metrics.}
We comprehensively evaluate {\tt AdaLog} using three key metrics. 
\textit{(i) Compression Ratio}: We measure the behavior log size reduction achieved by {\tt AdaLog} to quantify storage efficiency, expressed as the percentage decrease in log size compared to the original behavior log.
\textit{(ii) Feature Computation Time}:  To assess the impact of {\tt AdaLog} on ML inference speed, we measure the wall-clock time required to compute features for model inference with and without {\tt AdaLog}. This metric is crucial for verifying that the storage optimizations do not negatively affect real-time inference performance on mobile devices.
\textit{(iii) System Overhead}: We measure the execution time and peak memory usage when {\tt AdaLog} is invoked. This provides insight into the system's efficiency and its suitability for deployment on resource-constrained devices.

\subsection{Overall Performance}
\label{sec: overall performance}
We start with measuring the overall performance of {\tt AdaLog} across diverse application scenarios and mobile users.
\begin{figure*}
    \vspace{-0.2cm}
    \includegraphics[width=0.32\linewidth]{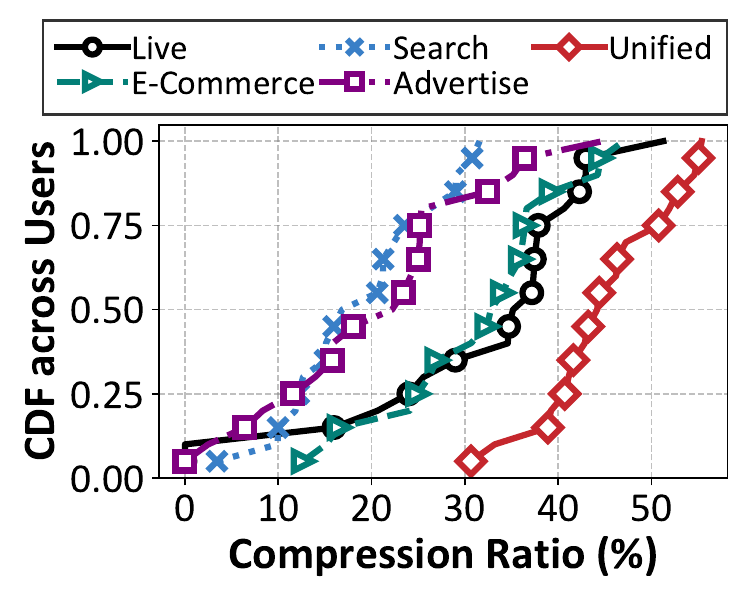}
    \includegraphics[width=0.32\linewidth]{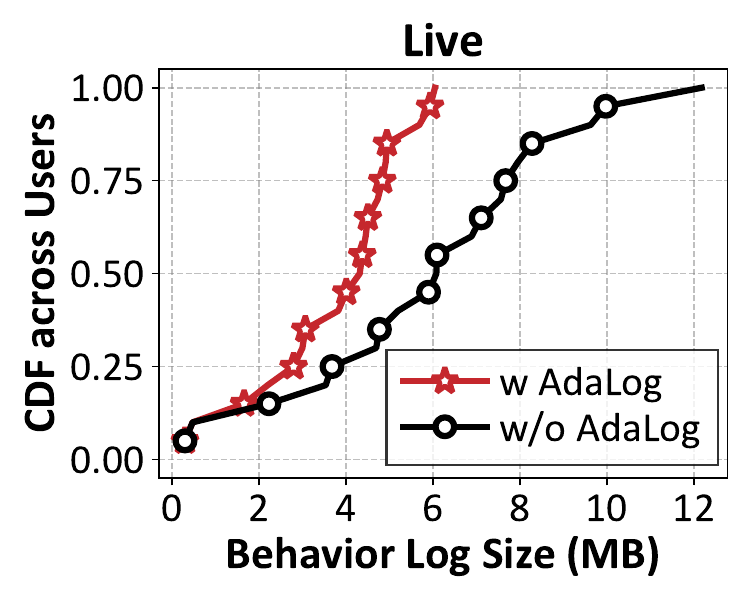}
    \includegraphics[width=0.32\linewidth]{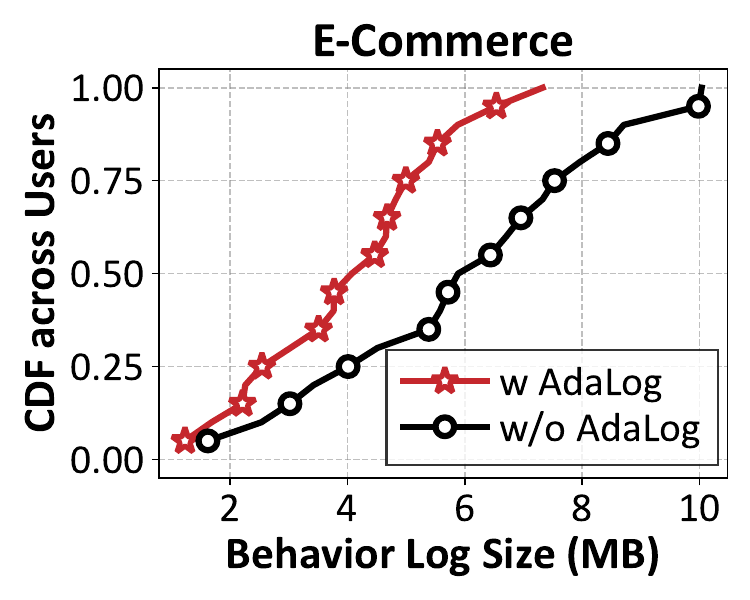}
    \includegraphics[width=0.32\linewidth]{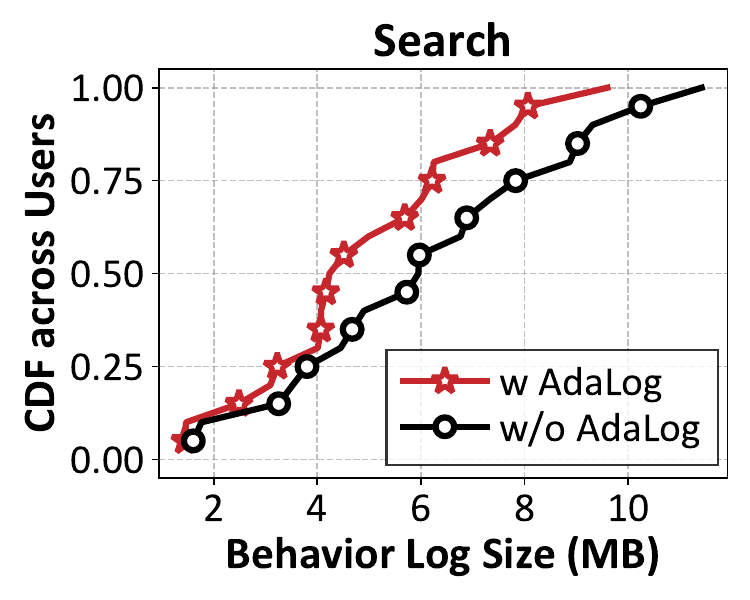}
    \includegraphics[width=0.32\linewidth]{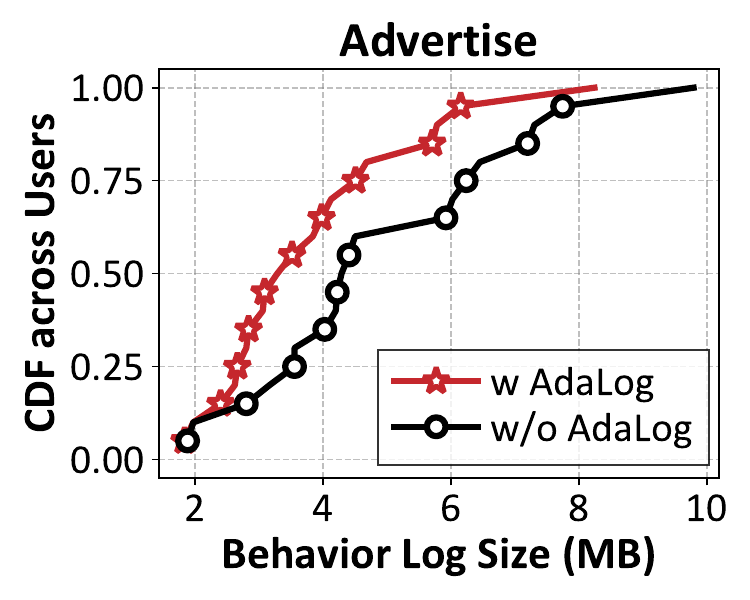}
    \includegraphics[width=0.32\linewidth]{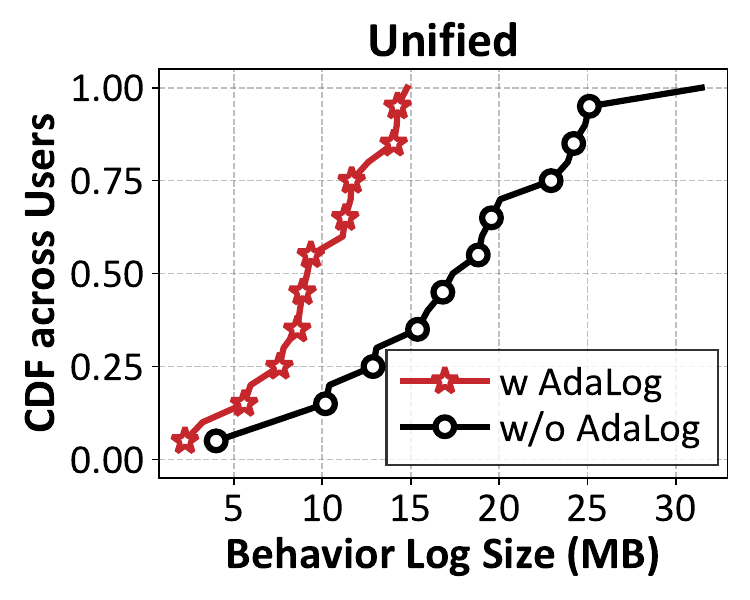}
    \caption{Distributions of compression ratio and behavior log size across users with and without {\tt AdaLog}. }
    \Description{Distributions of compression ratio and behavior log size across users with and without {\tt AdaLog}. 
    The absolute log sizes are on the order of tens of MBs, as experiments are based on 14 days of user data due to enterprise's limitations on online evaluation. In practical settings, the size would be significantly larger as existing features can require time windows up to 6 months, i.e., $12\times$ of currently presented sizes.}
    \label{fig: overall storage efficiency}
\end{figure*}

\textbf{{\tt AdaLog} significantly reduces behavior log storage overhead.}
Figure \ref{fig: overall storage efficiency} presents the distribution of {\tt AdaLog}'s compression ratio across different users in various application domains. 
Compared to the industry-standard behavior log design, {\tt AdaLog} achieves substantial reductions in storage consumption, with an average compression ratio of $35.1\%$ for live streaming domain, $32.8\%$ for e-commerce, $18.9\%$ for search and $23.4\%$ for advertisement. In the unified application case, where cross-scenario optimization is possible, {\tt AdaLog} achieves an impressive 44\% reduction in behavior log size for an average user.
This reduction translate to a $1.82\times$ increase in the numbers of on-device ML models (or ML-powered application services) that can be supported under the same storage cost. 
Figure \ref{fig: overall storage efficiency} further illustrates how {\tt AdaLog} impacts behavior log size distribution across users. Notably, users with larger behavior logs benefit the most, as they typically exhibit higher application usage and generate a greater volume of hot behaviors. These behaviors are often necessitated by massive features of on-device ML models and lead to high redundancy. This trend highlights the potential of {\tt AdaLog} in reducing more storage space for active users, ultimately improving the number of daily active users and application revenue.

\begin{figure*}
    \includegraphics[width=0.195\linewidth]{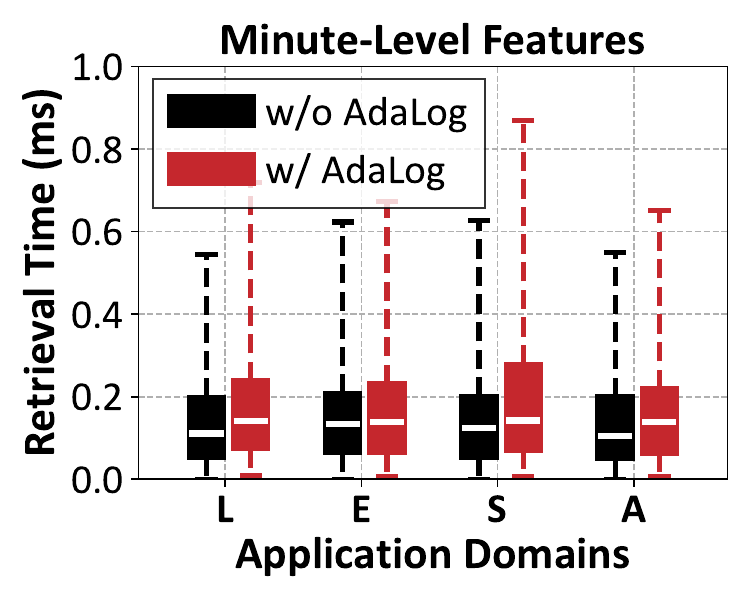}
    \includegraphics[width=0.195\linewidth]{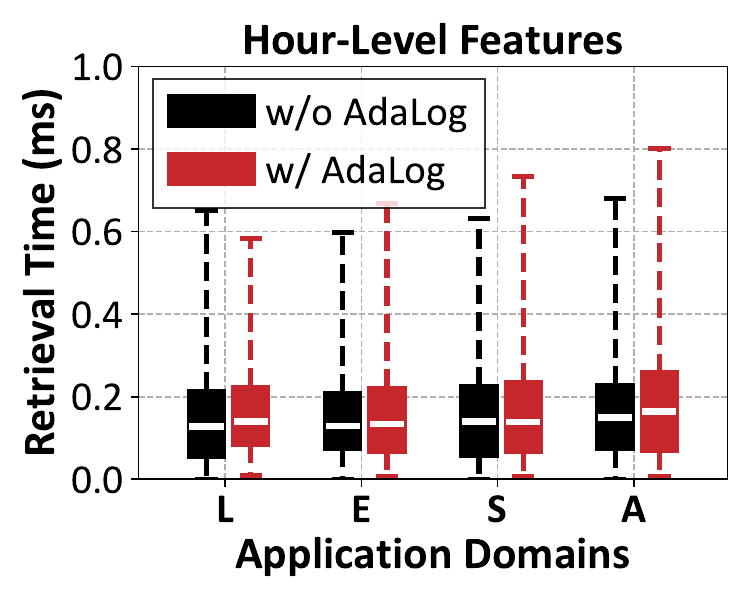}
    \includegraphics[width=0.195\linewidth]{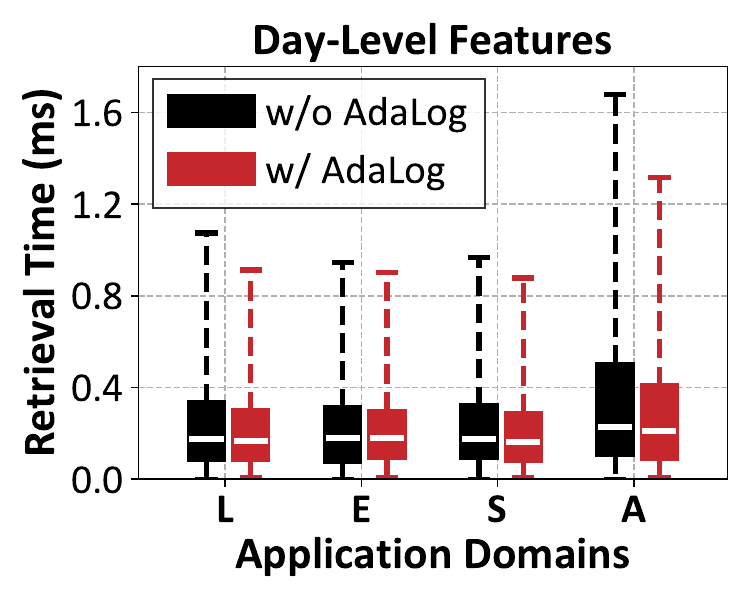}
    \includegraphics[width=0.195\linewidth]{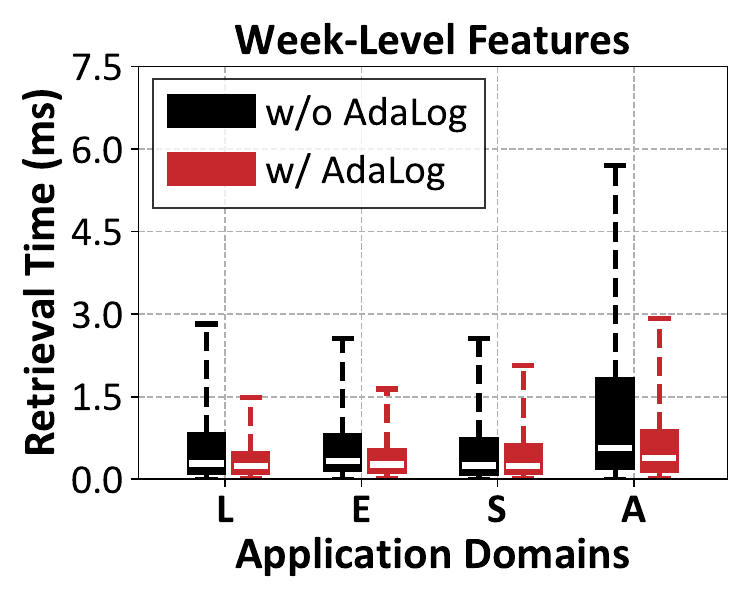}
    \includegraphics[width=0.195\linewidth]{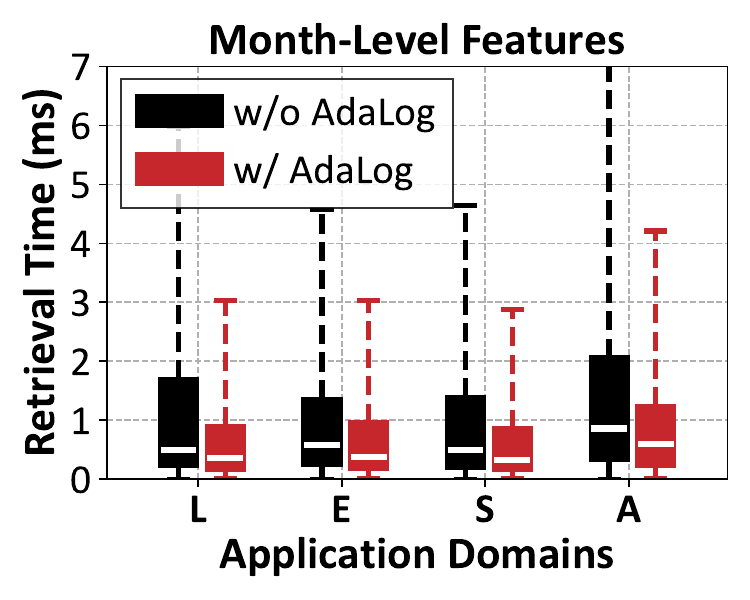}
    \caption{Wall-clock time across mobile users for computing different features with various time windows, where L, E, S, A represent scenarios of live streaming, e-commerce, search and advertisement.}
    \Description{Wall-clock time across mobile users for computing different features with various time windows, where L, E, S, A represent scenarios of live streaming, e-commerce, search and advertisement.}
    \label{fig: overall retrieve latency}
\end{figure*}

\textbf{{\tt AdaLog} preserves or improves feature computation speed for model inferences.}
A critical concern for storage optimization is its impact on real-time feature computation for on-device ML inference. Since {\tt AdaLog} introduces a one-time overhead by loading attribute mappings from a stored configuration file into memory, we evaluate its effect on feature computation latency. 
Figure \ref{fig: overall retrieve latency} presents the wall-clock time required to compute various features across mobile users, which are categorized by the time window of behaviors considered by each feature.
We observe that {\tt AdaLog} maintains near-identical computation speed for short-period features (minute and hour-level) while consistently improving retrieval efficiency for long-period features.
This improvement stems from two key factors. 
\textit{(i) Amortized Overhead}: The millisecond-level cost of loading attribute mappings is spread across multiple features during each on-device model inference, reducing its impact; 
\textit{2) Reduced Redundancy}: By reducing the number of event rows in behavior log files, {\tt AdaLog} speeds up database retrieval operations.

\begin{figure}
    \centering
    \includegraphics[width=0.9\linewidth]{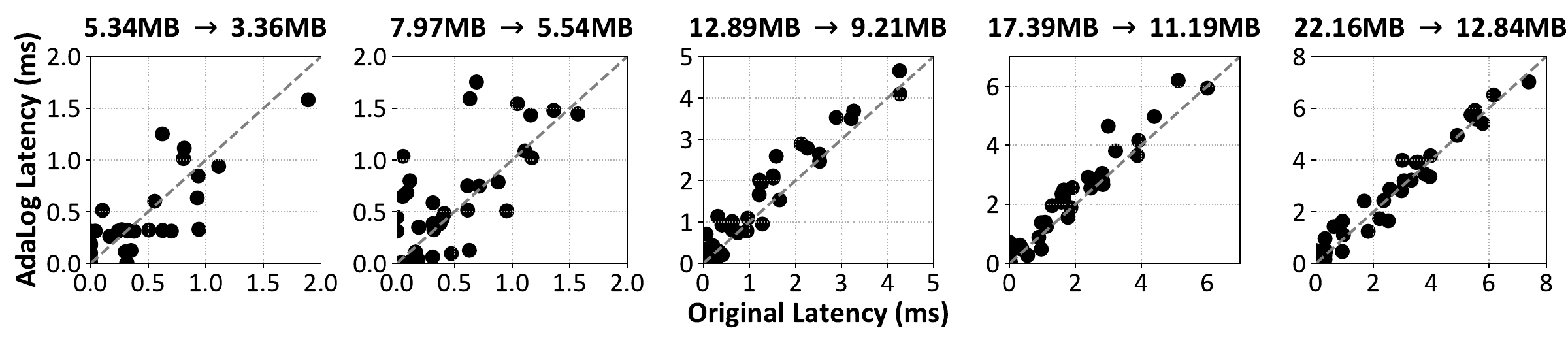}
    \caption{
    		We compare the data retrieval latency during on-device model inferences when using and not using {\tt AdaLog} system. Each subfigure represents a user with a distinct behavior log size. Within each subfigure, every data point represents the measured latency for a specific on-device ML model.
    }
    \label{fig: comparison of retrieve latency}
\end{figure}

	To thoroughly measure the impact of AdaLog on on-device model inferences, we collected behavior logs with diverse sizes from various users ranging from 5.34 MB to 22.16 MB, re-structured them with and without AdaLog, and measured the time required for each on-device model to retrieve its necessary data on our testing device iPhone 13 Pro. 
In Figure \ref{fig: comparison of retrieve latency}, we visualize the retrieval latency between original industry-standard log design and AdaLog system across 5 representative users' data. 
Overall, we observe that the data retrieval latency of AdaLog is consistent with the original industry-standard logging system. This confirms our goal that AdaLog achieves storage reduction without imposing a penalty on model inference speed.
For smaller behavior logs, AdaLog's performance exhibits instability and large variance, sometimes appearing inferior to the original log design. This is because the data retrieval latency is naturally very short for these small logs, making the measurement highly sensitive to transient device hardware states. In contrast, for larger logs, AdaLog maintains a stable and comparable retrieval speed to the original, unoptimized design, validating the effectiveness of our optimized storage structure.
As a result, {\tt AdaLog} effectively reduces storage overhead without compromising the responsiveness or personalization of ML-powered application services.

\begin{figure}
    \centering
    \subfigure[Time overhead of overall system and each key stage.]{
        \includegraphics[width=0.48\linewidth]{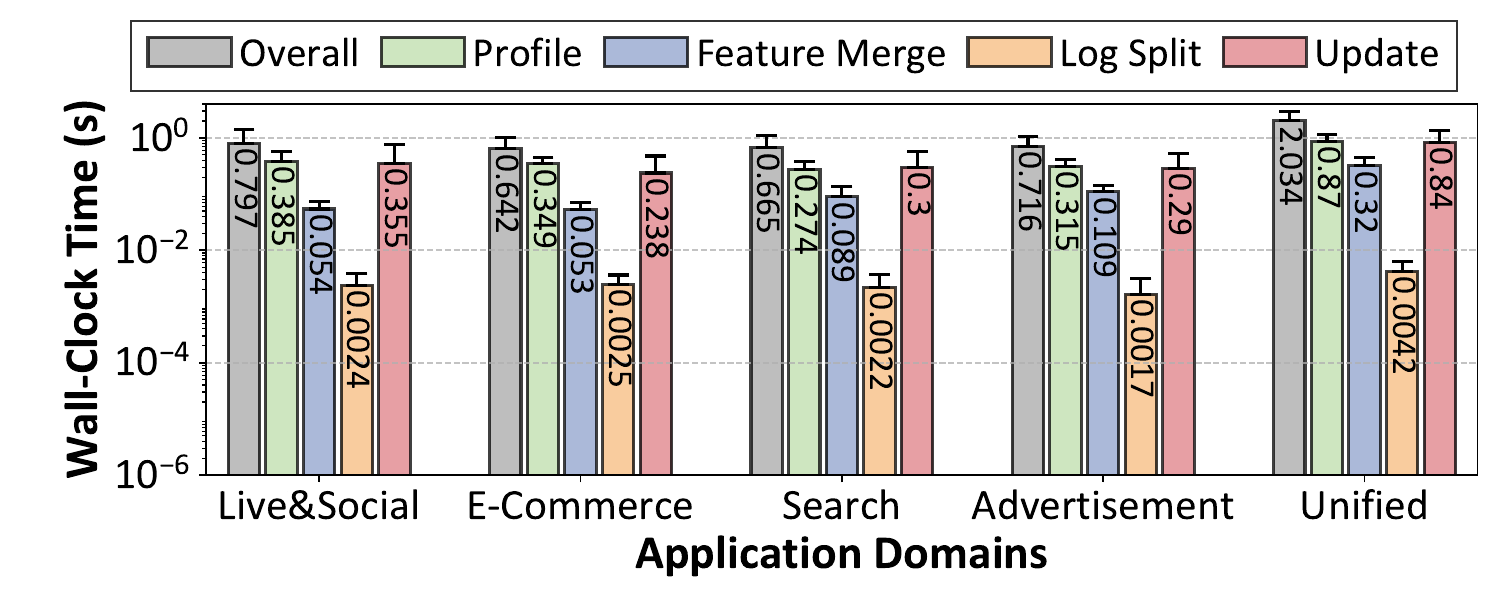}
        \label{fig: time overhead}
    }
    \centering
    \subfigure[Peak memory footprint during each key stage.]{
        \centering
        \includegraphics[width=0.48\linewidth]{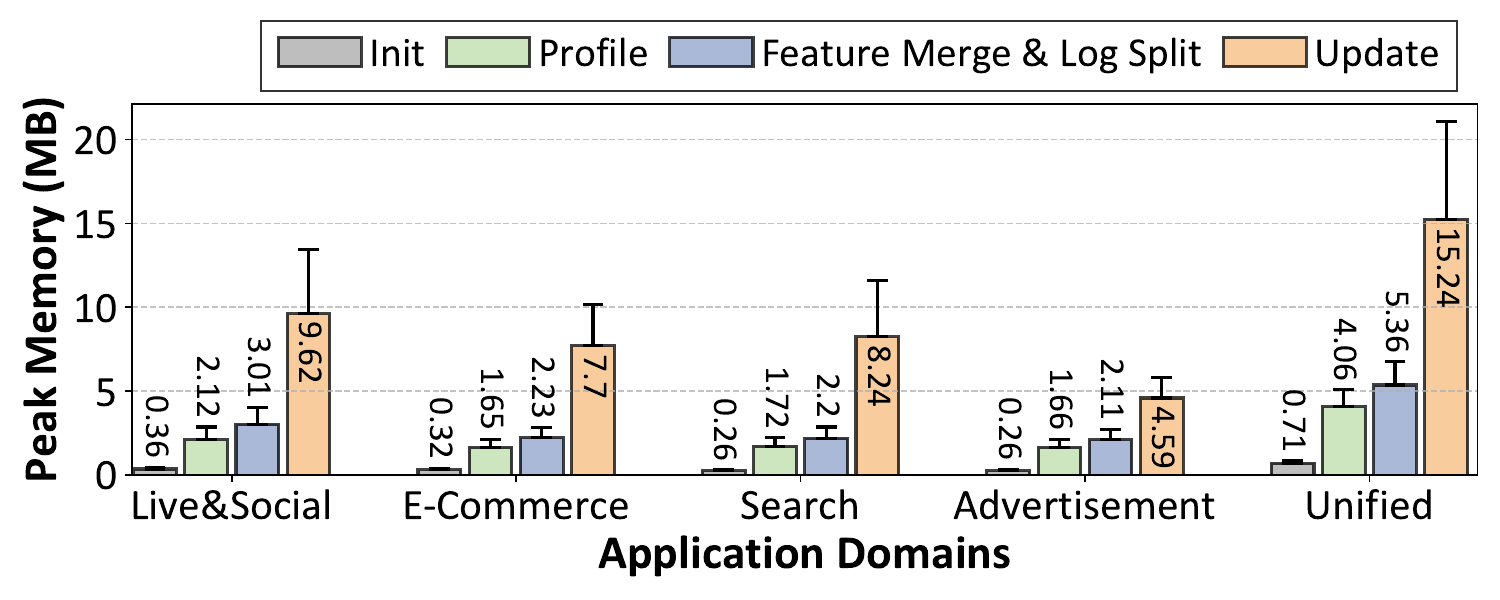}
        \label{fig: memory overhead}
    }
    \caption{System overhead of {\tt AdaLog} across mobile users for different mobile app domains.}
    \Description{System overhead of {\tt AdaLog} across mobile users for different mobile app domains.}
    \label{fig: system overhead}
\end{figure}

\textbf{{\tt AdaLog} introduces minimal system overhead.}
To ensure real-world deployment feasibility, we analyze {\tt AdaLog}’s execution overhead on mobile devices. Figure \ref{fig: system overhead} provides a breakdown of both time and peak memory consumption during {\tt AdaLog}'s daily execution. 

\noindent $\bullet$ \textit{Execution Time}: 
On average, {\tt AdaLog} completes its optimization process in 2.03 seconds for the unified case, with wall-clock time ranging from 0.642 to 0.797 seconds across different users and application scenarios.
As shown in Figure \ref{fig: time overhead}, 41\% of the time is spent on the incremental update process, 42\% is used for profiling necessary information, while only 17\%($0.05\!\sim\!0.32$s) is consumed for generating optimal configurations via feature merging and log splitting. 
The remarkably low latency is enabled by: 
(i) The low complexity of our proposed hierarchical merging algorithm and simple attribute-count-based splitting strategy, and (ii) System-level optimizations that conduct hierarchical merging algorithms for different behaviors in parallel.

\noindent $\bullet$ \textit{Memory Usage}:
Figure \ref{fig: memory overhead} illustrates the memory consumption during different processing stages. Peak memory usage occurs during the incremental update process, averaging 4.59-15.24MB across application scenarios. This is primarily due to the need to load and process event rows affected by configuration changes, typically involving hundreds of rows. 
A secondary memory peak arises during the feature merging and log splitting stage, which incurs an 2.2-5.36MB footprint due to constructing graphs for each behavior's related features.
Overall, {\tt AdaLog} maintains a memory footprint below 20MB, making it lightweight and well-suited for modern mobile devices.

\subsection{Component-Wise Analysis}
\label{sec: component-wise analysis}
To further validate the effectiveness of each key design in {\tt AdaLog}, we implement multiple modified versions on 20 voluntary mobile users in the unified application case.

\begin{figure}
    \begin{minipage}[b]{0.52\linewidth}
        \centering
        \includegraphics[width=\linewidth]{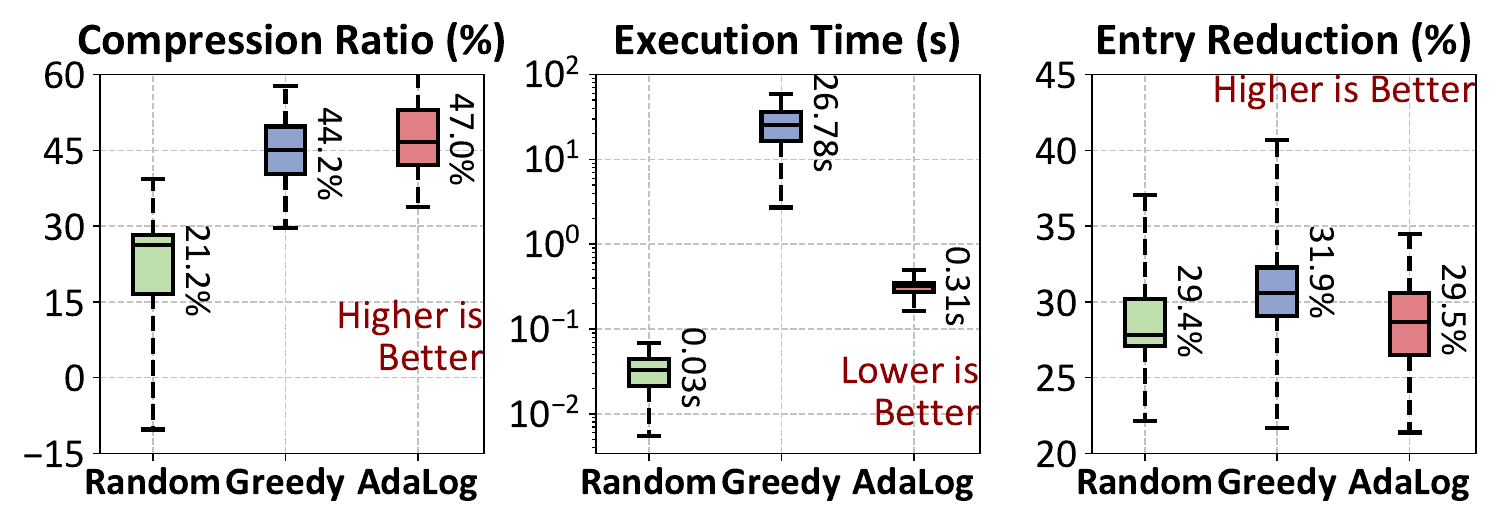}
        \caption{Effect of hierarchical merging algorithm in balancing storage reduction and efficiency.}
        \Description{Effect of hierarchical merging algorithm in balancing storage reduction and efficiency.}
        \label{fig: ablation of feature merging}
    \end{minipage}
    \ \ 
    \begin{minipage}[b]{0.46\linewidth}
        \centering
        \includegraphics[width=\linewidth]{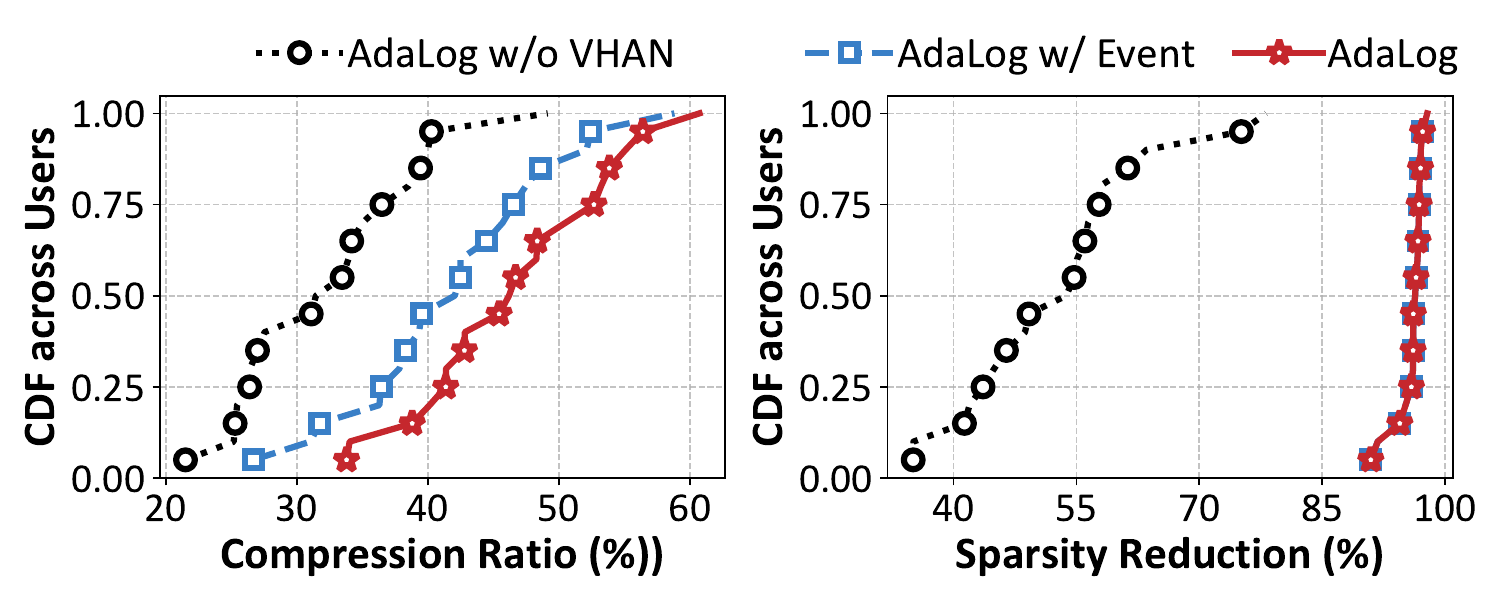}
        \caption{Effect of VHAN in reducing storage overhead and minimizing storage sparsity.}
        \Description{Effect of VHAN in reducing storage overhead and minimizing storage sparsity.}
        \label{fig: ablation of log splitting}
    \end{minipage}
\end{figure}

\textbf{Feature-Level Data Merging.}
A core innovation in {\tt AdaLog} is hierarchical merging algorithm, which balances data storage reduction, index size increase and computational efficiency. To evaluate its significance, we compare it against two alternative strategies:
\textit{(i) Random}: Features are randomly grouped for data merging, with performance averaged over $10$ trials;  
\textit{(ii) Greedy}: Features are grouped using a classic greedy algorithm~\cite{besser2017greedy} that prioritizes hyperedges (i.e., feature groups) with the highest weights (i.e., storage reduction) when solving the maximum weighted matching problem.
Figure \ref{fig: ablation of feature merging} presents a comparative analysis of compression ratio, execution time, and redundancy elimination across these methods. Our findings reveal two critical insights.
(i) {\tt AdaLog} achieves the optimal trade-off between compression and efficiency. While \textit{Greedy} approach achieves a compression ration comparable to {\tt AdaLog}, it suffers from 86$\times$ higher execution time due to the exponential number of hyperedges for given features. Conversely, \textit{Random} runs faster but performs $26\%$ worse in compression on average. 
(ii) Despite both \textit{Random} and \textit{Greedy} remove a similar number of redundant event rows, their practical storage reduction is lower. In some cases, \textit{Random} even increases storage costs due to the increased index overhead.
This highlights the necessity of hierarchical merging algorithm in ensuring maximal compression ratio without excessive computational burden.

We further analyze the specific scenarios where our pairwise hierarchical merging strategy might be practically sub-optimal. Our hierarchical pairwise algorithm operates by greedily merging the two feature groups that yield the largest potential storage reduction in each step. This method is near-perfect when the main source of redundancy is found between simple pairs of features. However, the algorithm's performance could theoretically degrade if the absolute maximum redundancy were only achievable by merging an odd number of groups simultaneously (e.g., three, five, or seven). However, the conditions for this theoretical sub-optimality are demonstrably extremely rare in real-world behavior logs. Our empirical analysis confirms that features rarely align in the complex, odd groupings that would challenge our algorithm. Across all testing users, we found that only up to 3 out of 100+ user behaviors involve 3, 5, or 7 features sharing highly redundant event rows. This means the maximal gain from optimally merging these rare high-order groups would contribute negligible storage reduction to the overall behavior log.

\textbf{Behavior-Level Log Splitting.}
Next, we assess the role of virtually hashed attribute name (VHAN) design in reducing storage sparsity and overall footprint. To quantify its impact, we compare to two modified versions of {\tt AdaLog}: 
(1) \textit{AdaLog w/o VHAN}: Uses attribute-count-based log splitting but disables VHAN, storing physical attribute names instead of virtual IDs.
(2) \textit{AdaLog w/ Event}: Groups behavior events by shared attribute sets, ensuring that all event rows within a single behavior log file have the same relevant attributes.
Figure \ref{fig: ablation of log splitting} illustrates the compression ratios and storage sparsity reductions across users. 
We notice that without VHAN, compression ratio decreases by 14\% and storage sparsity increases by 35\%. This highlights VHAN's effectiveness in consolidating diverse attributes into a single virtual attribute representation, eliminating null values.
Also, the \textit{AdaLog w/ Event} method achieves storage sparsity reduction comparable to VHAN but suffers from 8\% lower compression performance, caused by the substantial metadata overhead from managing hundreds of fragmented database files.

\textbf{Incremental Update Mechanism.}
Finally, we analyze the system efficiency of {\tt AdaLog}'s incremental update mechanism in adapting outdated behavior log files to new configurations over time. We compare {\tt AdaLog} with the \textit{reconstruction} method, which rebuilds behavior log files from scratch.
Figure \ref{fig: ablation of incremental update} shows the time and memory footprints for both methods over a two-week period. Our analysis reveals that as event rows accumulate, full reconstruction leads to linearly increasing execution time and memory consumption, whereas {\tt AdaLog}’s incremental update mechanism remains stable.
Compared to full reconstruction, incremental updates achieve a $3.4\times$ to $8.1\times$ speedup and a $2.2\times$ to $2.5\times$ reduction in memory consumption. By efficiently updating only event rows affected by configuration updates, {\tt AdaLog} prevents excessive resource consumption, ensuring minimal interference with other important on-device applications.

\subsection{Sensitivity Analysis}
\label{sec: sensitivity analysis}
Given that {\tt AdaLog} involves few hyperparameters and avoids trial-and-error processes, we focus on evaluating its performance under environmental factors such as timeline and model numbers. 

\begin{figure}
    \begin{minipage}[b]{0.485\linewidth}
        \centering
        \subfigure[Execution time.]{
            \includegraphics[width=0.46\linewidth]{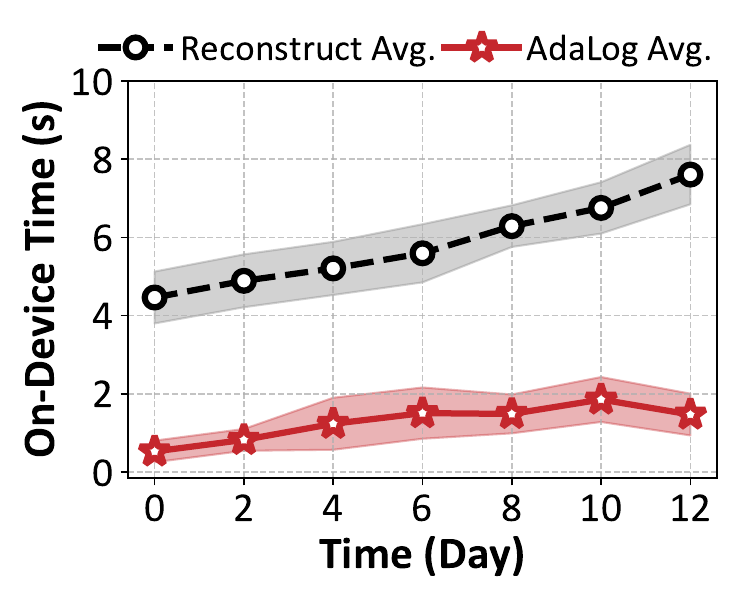}
        }
        \subfigure[Peak memory.]{
            \includegraphics[width=0.46\linewidth]{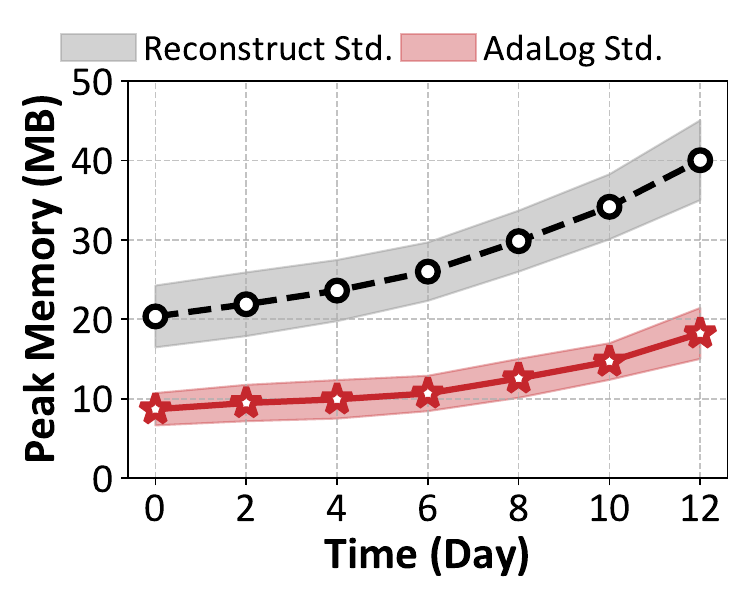}
        }
        \caption{System efficiency of incremental updates.}
        \Description{System efficiency of incremental updates.}
        \label{fig: ablation of incremental update}
    \end{minipage}
    \begin{minipage}[b]{0.485\linewidth}
        \centering
        \subfigure[Impact of time.]{
            \includegraphics[width=0.46\linewidth]{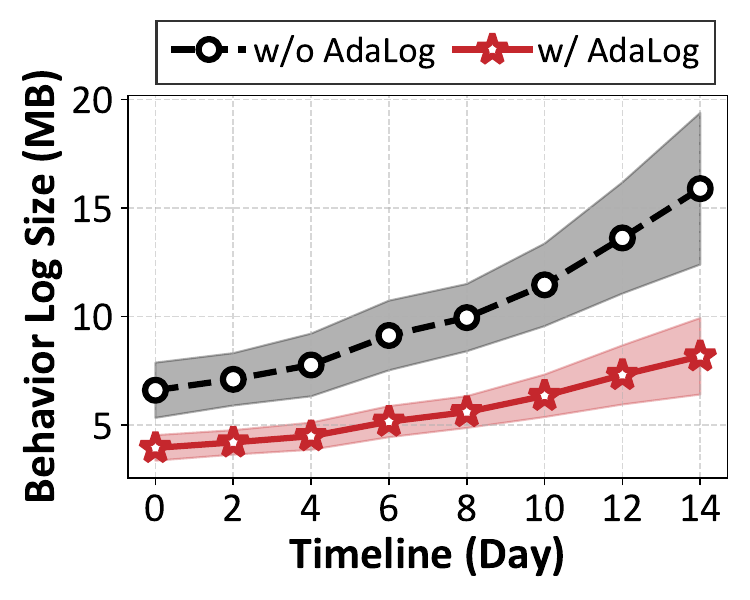}
            \label{fig: impact of time}
        }
        \subfigure[Impact of ML models.]{
            \includegraphics[width=0.46\linewidth]{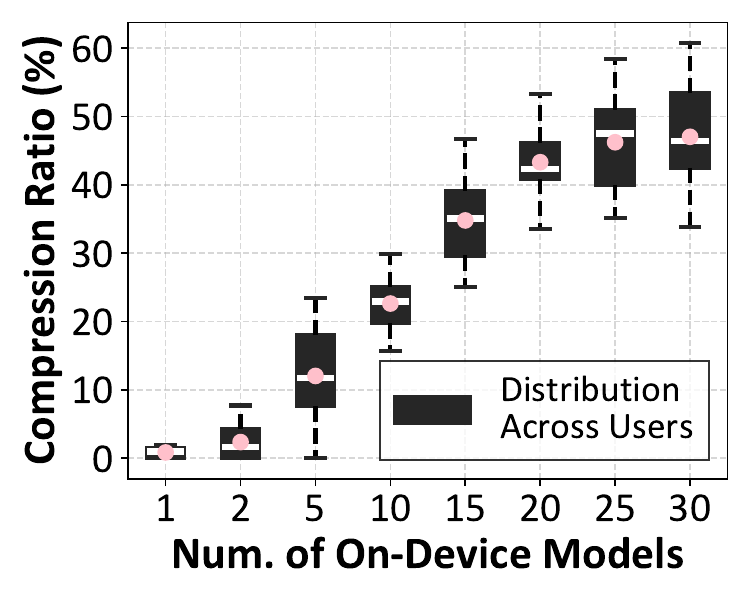}
            \label{fig: impact of model}
        }
        \caption{Sensitivity analysis on natural factors.}
        \Description{Sensitivity analysis on natural factors.}
    \end{minipage}
\end{figure}

\textbf{Impact of Time.}
Figure \ref{fig: impact of time} shows the evolution of behavior log file size over a 14-day period, which includes a public holiday.  
We observe that behavior log size of the industry-standard solution grows exponentially over time, attributed to two factors: 
(i) During holidays, users tend to interact more with the applications, resulting in more hot behaviors recorded in behavior log;
(ii) New ML models are deployed on mobile devices to test new application services, leading to more recorded event rows.
In contrast, {\tt AdaLog} exhibits a nearly linear growth in storage size, which is significantly slower than the original design. This trend is directly attributed to {\tt AdaLog}'s feature-level data merging technique, effectively reducing redundancy for hot behaviors. As a result, {\tt AdaLog} is more efficient in managing long-term storage, even during periods of increased user activity and model deployment.

\textbf{Impact of Model Number.}
Next, we explore how the performance of {\tt AdaLog} is affected by the number of ML models deployed within an application. For this analysis, we conducted experiments where testing users are forced to generate behavior log files for varying numbers of on-device models. 
Figure \ref{fig: impact of model} presents the relation between the number of models and the achieved compression ratio. As the number of models increases from 2 to 20, the compression ratio achieved by {\tt AdaLog} rises rapidly, from 2\% to 43\%. This demonstrates that {\tt AdaLog} scales effectively as more models are deployed on the device, efficiently reducing storage overhead.
After the number of models exceeds 25, the compression ratio gradually stabilizes at around 45\%. 
The primary reason is that with the increase of on-device models, newly deployed model is more likely to have input features with identical filtering conditions (i.e., the same FilterIDs) with previous models, implying that less redundant data is introduced. 

\begin{figure}
    \centering
    \includegraphics[width=0.4\linewidth]{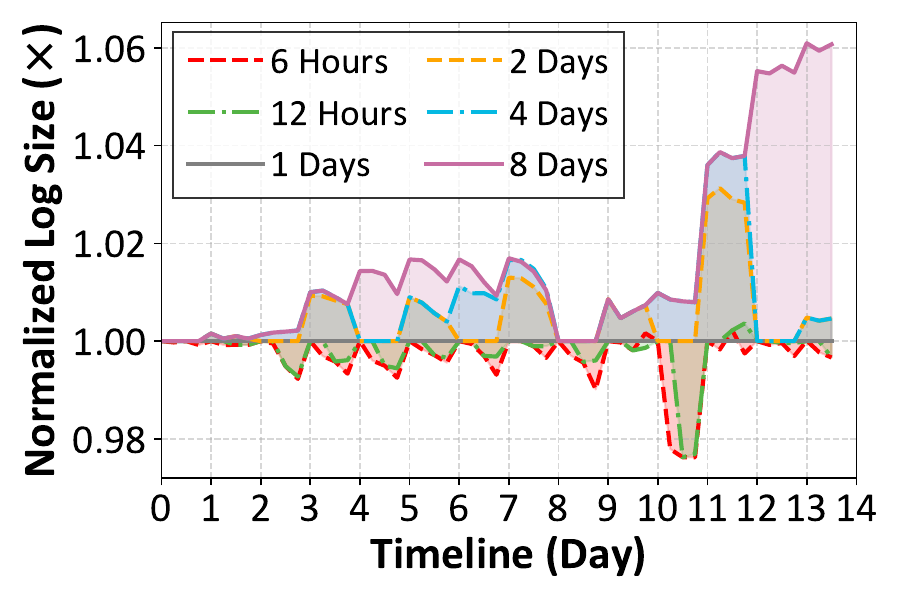}
    \caption{
    		Impact of update frequency on {\tt AdaLog}'s storage efficiency. For better comparison, behavior log sizes are normalized against the size achieved under a standard daily update schedule.
    }
    \label{fig: impact of update frequency}
\end{figure}
\textbf{Impact of Update Frequency.}
Further, we conduct experiments to analyze the impact of varying the log update frequency on {\tt AdaLog}'s performance.
We measured the average log size across testing users, varying the update intervals from 6 hours up to 8 days. As shown in Figure \ref{fig: impact of update frequency}, we plot the normalized behavior log sizes achieved by {\tt AdaLog} across various update frequencies. All sizes are normalized to the baseline value recorded when the log is updated daily for better visualization.
We notice that increasing the update frequency beyond one day (e.g., 6 hours and 12 hours) only results in a marginal improvement in compression, typically less than 2.2\%. Slowing down the frequency from one day to 4 and 8 days makes the behavior log gradually drifts from the optimal storage configuration, leading to a noticeable increase in log size of 4\%-6\%.
Based on these findings, we propose that the practical update process should be executed every 1 to 4 days, which can be dynamically controlled by factors such as user activity profile and available device hardware resources.

\section{Related Work}
\label{sec: Related Work}

\textbf{Resource-Efficient On-Device ML.}
Extensive research is conducted to optimize resource utilization for on-device ML through two main levels. 
System-level optimization directly improves resource efficiency, such as 
reducing memory footprint~\cite{DBLP:conf/mobicom/FangZ018, DBLP:conf/ipsn/LaneBGFJQK16, DBLP:conf/mobicom/WangDCLX21, DBLP:conf/mobisys/WangXJDY0HLL22, li2024flexnn}, 
enhancing computational hardware utilization~\cite{DBLP:conf/mobisys/CaoBB17, DBLP:conf/mobicom/WangDCLX21, DBLP:conf/mobicom/XuXWW0H0JL22, jia2022codl} and 
minimizing energy consumption~\cite{DBLP:conf/ipsn/LaneBGFJQK16, DBLP:conf/mobicom/FangZ018, kim2020autoscale}.
Model-level optimization compresses on-device models to make inferences more efficient, using techniques such as quantization~\cite{liu2018demand, kim2019mulayer, lin2024awq},     
pruning~\cite{DBLP:conf/mobicom/WenLZJYOZL23, shen2024fedconv, ma2020pconv, niu2020patdnn, lym2019prunetrain}, and sparsification~\cite{lym2019prunetrain, bhattacharya2016sparsification}.
Notably, most existing research overlooks storage as a critical resource. This is because they focused on traditional CV and NLP models that use static features, which do not require storing massive raw data for computing dynamic features.
Consequently, our work is complementary to them by improving storage efficiency for ML models in modern mobile applications.

\textbf{Input Filtering for Edge Computing.}
Input filtering aims to reduce unnecessary computation by filtering redundant or irrelevant inputs. Examples include raw data (e.g., undecoded packets~\cite{yuan2023packetgame}), entire input features (e.g., frames~\cite{chen2015glimpse, guo2018foggycache, DBLP:conf/sigcomm/LiPZWXN20, DBLP:conf/asplos/GuoH18}), and partial features (e.g., pixels~\cite{yuan2022infi, jiang2021flexible}). 
In this sense, our work can be seen as a new form of raw data filtering tailored for storage optimization rather than computation reduction, where we adaptively filter out redundant and null data in user behaviors recorded in application logs.

\textbf{Data Management for Mobile Apps.}
In the area of mobile data management, existing works mainly fall into two categories.
The first category optimizes data I/O costs during app usage, i.e., faster data writing and reading. Common optimization techniques include virtual page writing~\cite{oh2015sqlite}, SSD access optimization~\cite{song2023prism, liang2020itrim}, cache policy design~\cite{shen2020efficient}, etc. 
The second category targets storage optimization for time-series and sensor data, such as Apple HealthKit~\cite{north2016apple} and Samsung Health~\cite{jung2019development} aggregating and compressing IMU data collected from mobile device. 
We observe that few existing works consider the storage cost of user behavior \textit{events}, a commonly seen data format in mobile apps, because traditional apps simply upload them to cloud server for centralized storage. However, with the prevalence of intelligent app services supported by on-device ML models, keeping an on-device behavior log introduces an inevitable storage bottleneck, which is the focus of our work.

\textbf{Database Optimization Techniques.}
In the broader database community, redundancy and sparsity have traditionally been mitigated through structural and operational optimizations.
Columnar storage formats like Apache Parquet~\cite{ivanov2020impact} and ORC~\cite{apache} reduce redundancy by grouping similar values and applying compression methods to encode attributes into smaller symbols~\cite{muller2014adaptive, kuschewski2023btrblocks, abadi2006integrating}. 
Sparse indexing~\cite{lillibridge2009sparse} further optimizes storage by selectively indexing non-redundant rows.
However, these database-centric approaches focus on structural and system-level efficiencies without optimizing original storage content, and require customized modifications to the database backend. 
In contrast, {\tt AdaLog} tackles redundancy and sparsity at the data content layer and offers a more flexible and generalizable solution that can seamlessly integrate with existing database implementations and optimizations.
\section{Conclusion}
\label{section: conclusion}
In this work, we identified an overlooked behavior log storage bottleneck in ML-embedded mobile apps, which poses a significant challenge to development and broader deployment of on-device ML models.
To address this, we proposed {\tt AdaLog} system to enhance data storage efficiency by adaptively reducing redundancy and sparsity in behavior log, without compromising on-device inference accuracy or latency.
Extensive evaluations across real-world mobile users and application scenarios demonstrate that {\tt AdaLog} effectively reduces behavior log storage overhead with minimal system impact, providing a powerful and efficient data foundation for on-device ML deployment.

\begin{acks}
We sincerely thank our shepherd - Lishan Yang, and anonymous reviewers for their constructive comments and invaluable suggestions that helped improve this paper. 
This work was supported in part by National Key R\&D Program of China (No. 2023YFB4502400), in part by China NSF grant No. 62322206, 62132018, 62025204, U2268204, 62272307, 62372296. The opinions, findings, conclusions, and recommendations expressed in this paper are those of the authors and do not necessarily reflect the views of the funding agencies or the government.
\end{acks}

\bibliographystyle{ACM-Reference-Format}
\bibliography{main.bib}

\clearpage
\appendix
\section{Diverse User Behaviors}
\label{appendix user behavior}
We list the top 50 user behaviors in industrial mobile apps in the following 4 tables for justification, which are partially modified to satisfy confidential requirements.
\begin{table}[h]
    \begin{minipage}[t]{0.48\linewidth}
    	\centering
        \caption{Top 1-25 Behaviors.}
        \begin{tabular}{|c|c|c|c|}
        \hline
            0 & play\_time\\
            1 & video\_play\\
            2 & show\\
            3 & page\_show\\
            4 & homepage\_slide\_up\\
            5 & play\_session\\
            6 & live\_show\\
            7 & trending\_words\_show\\
            8 & video\_finish\\
            9 & click\_comment\\
            10 & video\_stop\\
            11 & stay\\
            12 & like\\
            13 & live\_window\_show\\
            14 & show\_product\\
            15 & ad\_gap\\\
            16 & othershow\\
            17 & enter\_room\_duration\\
            18 & live\_window\_duration\\
            19 & room\_not\_render\\
            20 & play\\
            21 & trending\_show\\
            22 & silence\_launch\_app\\
            23 & video\_pause\\
            24 & back\_quit\\
            \hline
        \end{tabular}
        \label{tab: top behaviors 1}
    \end{minipage}
    \begin{minipage}[t]{0.48\linewidth}
    	\centering
        \caption{Top 26-50 Behaviors.}
        \begin{tabular}{|c|c|c|c|}
        \hline
            25 & live\_duration\\
            26 & inner\_push\\
            27 & wormhole\_preview\\
            28 & enter\_personal\_detail\\
            29 & homepage\_notice\\
            30 & search\_result\_show\\
            31 & homepage\_slide\_down\\
            32 & performance\_monitor\\
            33 & homepage\_tab\_stay\_time\\
            34 & enter\_homepage\\
            35 & wormhole\_preview\_reuse\\
            36 & live\_play\\
            37 & enter\_homepage\_message\\
            38 & search\\
            39 & show\_product\_v2\\
            40 & product\_entrance\_show\\
            41 & adjust\_volume\\
            42 & product\_entrance\\
            43 & post\_comment\\
            44 & search\_result\_click\\
            45 & show\_card\\
            46 & share\\
            47 & enter\_tab\\
            48 & search\_click\\
            49 & click\_product\\
            \hline
        \end{tabular}
        \label{tab: top behaviors 2}
        \end{minipage}
\end{table}

\end{document}